\documentclass[twoside,11pt]{article}

\usepackage{jmlr2e}
\usepackage{xcolor,graphicx}
\usepackage{hyperref}
\usepackage{amsmath}
\usepackage{float}
\usepackage{enumerate}
\usepackage{booktabs}
\usepackage[utf8]{inputenc}
\usepackage[inline]{enumitem}
\usepackage{algorithm}
\usepackage[noend]{algpseudocode}

\ShortHeadings{Common-Description Learning}{Basem Elbarashy}
\firstpageno{1}

\begin{document} \thispagestyle{empty}

\title{Common-Description Learning: A Framework for Learning Algorithms and Generating Subproblems from Few Examples}

\author{\name Basem G. El-Barashy \email es-basem.abdelghafar1116@alexu.edu.eg \\
       \addr Department of Electrical Engineering\\
       Alexandria University\\
       Alexandria, 21533, Egypt
      }


\maketitle

\begin{abstract}
Current learning  algorithms face many difficulties in learning simple patterns and using them to learn more complex ones. They also require more examples than humans do to learn the same pattern, assuming no prior knowledge.
In this paper, a new learning framework is introduced that is called common-description learning (CDL). This framework has been tested on 32 small multi-task datasets, and the results show that it was able to learn complex algorithms from a few number of examples.
The final model is perfectly interpretable and its depth depends on the question. What is meant by depth here is that whenever needed, the model learns to break down the problem into simpler subproblems and solves them using previously learned models. 
Finally, we explain the capabilities of our framework in discovering complex relations in data and how it can help in improving language understanding in machines.
\end{abstract}


\section{Introduction}
In this paper, I introduce a new framework called common-description learning (CDL).
Common-description (CD) is a discrete and nonparametric model, and is designed to capture two basic relations that are extremely simple: equality and adjacency. An equality relation means that the model can check whether two variables are equal or not, while adjacency means that the model can move through a sequence only one step  at a time (forward or backward). For example, reading the third word in a sequence cannot be done without passing through the first and second  word. These simple operations, accompanied with the ability to push data to memory, are the main operations used by the CD model to solve problems or handle memory.

To test the CD model, I constructed 32 small multi-task datasets each containing no more than 16 training sequences. We consider various tasks including, but not limited to, addition of two one-digit numbers, addition of three one-digit numbers using what is learned from the previous task, counting, copying, reverse, distinguishing between short and long input sequences, categories, comparing digits (greater than or less than), one and two supporting facts, inductive reasoning, different kinds of relations, and more.

The main adopted principles in this work are based on the belief that an intelligent machine should be able to learn simple patterns from a few number of examples, and to use the learned models in learning more complex ones. It should also be able to interact with human users, and thus we need a communication algorithm to make teaching machines easier and we also need a technique for tuning hyperparameters that can be easily done by non-experts. 

A variety of recent works have tried to learn simple patterns (algorithms). For example, \citet{zaremba:2015} used an RNN-based controller, trained using Q-learning, that interacts with the environment through a set of interfaces selected manually for each task. However, they failed to find one controller suitable for all tasks, and the learned models overfit the length of input in some tasks. \citet{weston:2015} provided the community with a great dataset of 20 QA tasks (baby tasks) and limited the answers to a single word. They showed that standard MemNNs outperforms N-gram and LSTM but failed in a number of tasks. For each task they used 1000 questions for training, and 1000 for testing. Meanwhile, the largest dataset presented in this paper has only 16 training examples. \citet{joulin:2015} used a stack RNN to learn simple algorithmic patterns generated by different sequence generators. 
Neural Turing Machine \citep{graves:2014} used a modified version of LSTM to learn a set of tasks, such as copying and sorting data sequences, and did not overfit the length of the  training sequences. 
CDL is different from the prior work as it can learn from small datasets, has different flavor of dynamic deep architecture that is shaped by the input sequence, requires less supervision, can learn which training sequences to memorize and how to handle its memory in a simple way, and finally has successfully learned 25 multi-task datasets that vary in complexity with the same hyperparameter settings without overfitting.

Standard deep models always require large training data, which costs money for collecting and labeling it and more importantly leads to simpler models \citep{halevy:2009}. Thus, small datasets could be a good metric for comparing models and that is why we used small datasets in evaluating the proposed model. Developing powerful communication tools \citep{mikolov:2015} that allow interaction between human users and machine is necessary to figure out what a machine learns and what it misses, and is also a complement to  the previous idea. Meaning that if we used small datasets, then it is expected that the machine may learn something else that also solves the training data. In that case, if what it learned is interpretable, then we can provide it with counter examples that reject what it learned and by repeatedly doing this we end up with the right model, and that is how the 32 datasets presented in \nameref{AppendixA} were built.

CDL is implemented in C++ language , and two supporting tools written in python are provided to help in understanding, analyzing, and interacting with the final model.  One tool can be used to visualize the common-description model (CD), and the other to show how the final model solves the test sequences. These visualization  tools are important because understanding how a model solves a problem could help in discovering its weaknesses and correcting them. Full source code, visualization tools, and the datasets can be found at {\color{blue} \url{https://github.com/BasemElbarashy/CDL}}. The reader can also find some GIFs that show how the learned models solve the test sequences in several tasks and can also visualize any learned model from the 32 experiments.

\section{Dataset}
The dataset is divided into 32 small datasets presented in \nameref{AppendixA}, and each one is composed of two parts: training data  and test data. Training data are a set of $m$ sequences of words
\begin{equation}
	s_i = (x_i,y_i)  \nonumber
\end{equation}
where $x_i$ is the sequence of all words of the $i^{th}$ sequence except the last word, and $y_i$ is the last word. The objective of the final model is to predict the last word in each sequence in the training data. There are no assumptions in our model about the number of tasks to be learned in the small datasets or their relatedness. Tables \ref{tab:c} and \ref{tab:b} are examples of two datsets.

\begin{table}[H]
\begin{minipage}{.4\linewidth}
	\centering
\begin{tabular}{|l|}
	\hline
	Training data \\
	1 1 1 2 3 4 =  {\color{red} 2} \\
	1 1 5 6 = {\color{red} 5} \\
	1 1 1 1 7 8 = {\color{red} 7} \\
	1 1 1 1 1 4 3 2 7 8 = {\color{red} 4} \\
	\\
	\hline
	Test data \\
	1 9 = {\color{red} 9} \\
	1 1 1 1 1 1 1 1 10 = {\color{red} 10} \\
	\hline
\end{tabular}
	\caption{The CDL model has to learn that the answer is the number after the ones}\label{tab:c}
\end{minipage}%
\hfill
\begin{minipage}{.48\linewidth}
	\centering
	\begin{tabular}{|l|}
		\hline
		Training data \\	
		1 2 3 E a b = c = d = f g =  {\color{red} f}\\
		4 5 E b = c d = n g = t =  {\color{red} n}\\
		7 E a  =  m d = f = l =  {\color{red} m}\\
		1 2 3 4 5 6 E f s = = = = = a = g =  {\color{red} g}\\
		\\
		\hline
		Test data \\
		1 2 3 4 5 E = = f = f = f = y =  {\color{red} y}\\
		h o E = 1 = 2 = 4 = 5 6 7 = 8 =  {\color{red} 2}\\
		\hline
	\end{tabular} 
	\caption{A complex pattern that can be solved with CDL as  discussed in subsection \ref{Cycles}.}\label{tab:b}
\end{minipage} 
\end{table}

One of the most important advantages of CDL over other machine learning algorithms is that it does not require a big training dataset to learn some task. Moreover, few number of examples can be used to test the model, and evaluation will be based on the accuracy of prediction and the steps the model takes to solve the problem. Steps that can be easily observed from the visualization tools that show how the final model solves the test questions.

\section{Model}
Common-description model is the central part of CDL. By `common' I refer to the capability of this model to produce a description that is found to be applicable to a number of training sequences which makes it more likely to be generalized to test sequences. CD solves the problem in a sequence of steps and is built with small blocks during training called nodes; each node is supposed to do a very simple operation and then point to the next node. In this section, different types of nodes and CDs are explained in detail.

\subsection{Variables}
CD describes the pattern sequentially by defining variables, comparing them, and using them to break down the question into simpler subproblems. Variables are defined over question, memory, or the generated subproblem. Variables value could be: 
\begin{itemize}
	\item Position: S and E are predefined variables; S is a virtual word located before the input sequence $x_i$, and E is a virtual word located after the last word in the sequence.
	\item Value: all values mentioned in the training data are saved as predefined variables, and also the variables are defined in solve node (see subsection \ref{sec:nodes}).
	\item Position and value: the variable which is defined by the assignment node and carries the position and value of some word in a sequence.
\end{itemize}
Z is the last variable defined in CD that carries its output.

\subsection{Nodes} \label{sec:nodes}
Having small bricks gives more freedom in shaping your construction. Therefore, nodes are designed to be as small as possible. Meaning that you cannot do what a node does in more than one step. Table \ref{tab:nodes} illustrates the different node types in the CD model.
\begin{table}[!htb]
	\begin{minipage}{\linewidth}

		\centering
		\begin{tabular}{|l|l|}
			\hline
			Assignment & defines a new variable that is located directly after or before another variable.\\
					   & If the new defined variable has the same position of S or E, then the next  \\
					   & node will be the sink node and the CD fails\\
			\hline
			Conditional & checks whether two variables are equal. It has two outgoing edges: true edge \\& which points to the next node if the comparison is true, and false edge which \\& points to the next node if the comparison is false  \\
			\hline
			Source& points to the first node in CD to apply on the input sequence\\
			\hline
			Sink& last node in CD		\\
			\hline
			Z  & defines Z variable, the output of the CD\\
			\hline
			Push  & pushes one word in the subproblem queue; CD can generate subproblems  \\& using more than the push node\\
			\hline
		    Solve  & solves the subproblem using the final model MGICD (see subsection \ref{sec:MGICD}),\\& and defines new variable carrying the answer	\\
			\hline
		    
		\end{tabular}
		\caption{Node Types}\label{tab:nodes}		
	\end{minipage}%
\end{table}

\subsection{Common Description (CD)}
CD is a discrete model that consists of nodes and flow edges, and defines the pattern in sequential steps. Having a training dataset, we start learning it by generating CDs that solve the first sequence and the same for other sequences. Then, some of them will be combined to generate the final model.
CD$_j(x)$ is the solution of $Z$ variable after applying the CD$_j$ on sequence $x$. However, the output of CD is undefined only if a SINK node is reached before the Z node. We use a characteristic vector $p_j \in [R,W,U]^m$ to describe the answers of CD$_j$ compared with the true answer of the training sequences; `R' means right answer, `W' means wrong answer, and `U'  means undefined.

\[ p_{ji} =
\begin{cases}
	R      & \quad CD_j(x_i) = y_i\\
	W      & \quad CD_j(x_i) \neq y_i\\
	U      & \quad CD_j(x_i) = \text{undefined}\\
\end{cases}
\]

We start by a very simple dataset to show how CD can describe a pattern and why combining them can be useful. The dataset in Table \ref{tab:a} requires the model to learn whether the answer is `M' or `G'.

\begin{table}[H]
	\begin{minipage}{\linewidth}
		\centering
		\begin{tabular}{|l|l|}
			\hline
			Training data & Test data\\
			\hline
			A B {\color{red} M} & A K {\color{red} M} \\
			A C {\color{red} M} & O F {\color{red} G} \\
			A D {\color{red} M} &\\ 			
			A F {\color{red} G} &\\
			H F {\color{red} G} &\\
			\hline
		\end{tabular}
		\caption{Simple dataset consisting of only five training sequences used to demonstrate how CD learning works. The answers are colored in red. }\label{tab:a}		
	\end{minipage}%
\end{table}

\begin{figure}[H]
	\includegraphics[width=8cm]{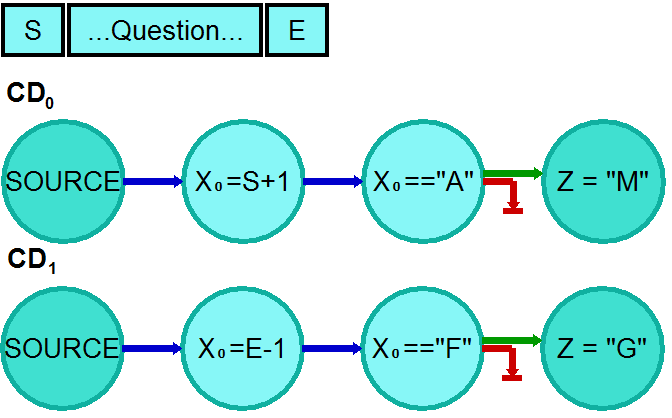}
	\centering
	\caption{CDs that could be learned from the dataset in Table \ref{tab:a}. By applying CD$_0$ on the first training sequence in Table \ref{tab:a}, the Source node points to the assignment node which defines $x_0$ as the first word after S. The assignment node points the conditional node that compares $x_0$ and `A'. If they are equal, the next node will be the Z node and the output will be `M'. Otherwise, the next node will be a Sink node and the output will be undefined. The sink node is represented by a small dash, the true edge of the conditional node is green, and the false edge is red which is connected to a sink node in both CDs.} \label{fig:a}
\end{figure}

The first thing to do is to generate candidate CDs that solve each individual sequence.
Figure \ref{fig:a} shows two CDs learned from the first and fourth sequences. The assignment node is called positive when it defines a new variable after another variable in the input sequence, like the assignment node in CD$_0$, and is called negative assignment if the new variable is defined before another one like the assignment node in CD$_1$.

For those CDs, the characteristic vectors become
\begin{align}
	p_0 & = (R,R,R,W,U)   \nonumber\\
	p_1 & = (U,U,U,R,R)   \nonumber
\end{align}
We got $p_0$ by applying CD$_0$ in Figure \ref{fig:a} on all training sequences. For the first three sequences, the answer will be `M' because the first word is `A' and that is the right answer. But since the first word in the fourth sequence is also `A', the answer of  CD$_0$ will be `M' which is wrong. The first word in the fifth sequence is not `A' and thus the output of CD$_0$ will be  undefined.

\subsection{Positive and Negative Descriptions} \label{Positive and Negative Descriptions}
To describe anything in the world, we either use features that it has (positive description), or features that it does not have (negative descriptions), or both. To describe your car by its color you could say ``My car is red", or you could say ``My car is not blue", ``My car is not yellow" ...etc. It is easy to see that the number of negative descriptions is extremely higher than that of positive ones; if the color value is continuous, then there is only one positive description and an infinite number of negative ones. However, if there are only two cars in the garage (red and blue), then the description ``My car is not yellow" is meaningless and there is only one negative description that can be useful ``My car is not blue" which is the negation of the positive description ``My car is blue" and hence refers to the red car.

\begin{figure}[H]
	\includegraphics[width=8cm]{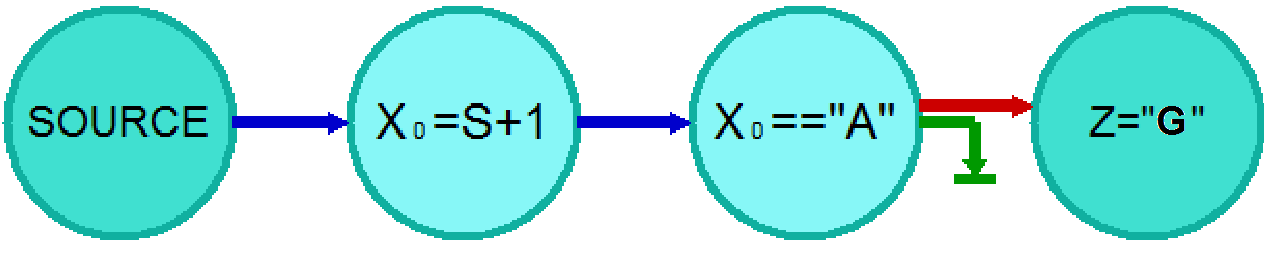}
	\centering
	\caption{This CD outputs `G' if the first word is \emph{not} `A'. Although it can solve the fifth training sequence in Table \ref{tab:a}, it will not be accepted in our framework because it describes a feature that the input sequence does not have (the true outgoing edge is connected to a sink node).} \label{fig:b}
\end{figure}

One of the most important ideas in learning CDs, which efficiently reduces the number of negative descriptions, is that we learn only positive common descriptions (PCD) of sequences and  use them as positive or negative descriptions of $x_i$ depending on the value of $p_{ji}$. For example, the CD in Figure \ref{fig:b} that describes a negative feature will not be learned. Therefore, False edges (colored in red) of conditional nodes must be connected to sink nodes except in cycles which are covered in Section  \ref{Cycles}.

\subsection{Hybrid CD} \label{HybridCD}
Hybrid CD (HCD) is a combination of two or more CDs in which the first one is considered a positive description and the rest are negative descriptions. Hence, the answer of the HCD is the answer of its PCD on the condition that all its NCDs are undefined when applied on a sequence. 
\[ HCD(x) =
\begin{cases}
PCD(x)    &  \quad \text{if } NCD(x) = U\\
U  &  \quad \text{Otherwise}\\
\end{cases}
\]
Back to our example, CD$_0$ is a good model for the first three sequences but it gives wrong answer when applied on the fourth sequence. We can build a HCD$_2$ considering CD$_0$ as the positive part and CD$_1$ as the negative part.
Now we get the new characteristic vector
\begin{equation}
p_2 = (R,R,R,U,U) \nonumber
\end{equation}	
$p_2$ takes the first three values of $p_0$ because CD1 is undefined on the first three sequences. The last two in $p_2$ are `U' because CD$_1$ is defined on the fourth and fifth sequences.

\subsection{Valid CD (VCD)}
CD$_j$ is valid if its characteristic vector $p_j$ contains no `W' and at least one `R'. A VCD can be PCD or HCD. In our example, we have two PCDs and one HCD, two of them are valid: CD$_1$ and HCD$_2$. Now we have two possible values in characteristic vectors: `R' and `U'. Thus, we can replace them by 1 and 0 respectively
\begin{align}
v_1 & =  (0,0,0,1,1)  \nonumber\\ 
v_2 & =  (1,1,1,0,0)  \nonumber
\end{align}

\subsection{Most General Integrated CDs (MGICD): The Final Model} \label{sec:MGICD}
Integrated CDs (ICD) are defined as the set of VCDs for which their characteristic vectors add up to the all-ones vector. Simply, ICD are a set of valid CDs that solve all sequences in the dataset. With enough capacity, it is guaranteed, and can be easily proved, that we can always find at least one ICD that solves all sequences
\footnote{If we have a sequence with n words, then we can always get a positive CD that defines n variables one after the other and checks whether all variables have the same values of that sequence or not. We can also get a  negative CD that only defines n+1 variables to ensure that we do not have other words in the input sequence. The resultant HCD is only defined when the input sequence is exactly the same as that sequence. With enough capacity and assuming there are no sequences in the dataset with the same $x$ and different $y$, we can do the same with each sequence in any dataset and the resultant HCDs are the ICD that solve all sequences in it.}, 
unless we have two sequences in a dataset with the same $x$ and different $y$ . The number of possible ICDs grows exponentially with the number of sequences in a dataset, which is one the challenges of CDL  discussed in detail in section \ref{learningMGICD}.

There are always a large number of integrated CDs that we can get but we need to find the most general one. Minimum description length principle \citep{grunwald:2007, barron:1998} is a general method for inductive inference that views learning as data compression. In brief, the more regularities we find, the more the data can be compressed, and learning models also try to find regularities in data. Therefore, description length criterion can be useful for model selection as it is in compression.

To apply that criterion we need first to define the concept of length in the context of CDL. Two candidates are memory space, identified by the number of nodes of the CD, and computation time, identified by the average number of steps taken by the CD to solve the sequences in training data. Experiments listed in section \ref{Experimental Results} shows that both are important but the former is much more important. Moreover, studying the relative importance between both criteria is required. Therefore, the number of nodes is the only criterion used in the experiments.

The criterion that I use is very simple, but it helped in finding very powerful models in different tasks. It is the number of nodes of the PCDs which are used to build the ICD. In our example, CD$_1$ and HCD$_2$ are ICD that can be selected to be our final model if their number of nodes is the smallest number we can get (10 nodes including sink nodes). However, the MGICD that we got in our experiment has 8 nodes because in CD$_0$ we do not need to check whether the first word is `A' or not. CD$_1$ is used twice in our model: once as a positive description, and the other as a  negative description in the hybrid CD but its nodes are counted once.

MGICD is the final model that we use to solve new questions by applying each one of its VCDs on the test sequence; it succeeds to give answer only in two cases:
\begin{enumerate}[label=\roman*),itemjoin={,\quad},itemsep=0pt]
	\item If all VCDs give `U' except one, then its output is our answer (more common).
	\item If more than one VCD does not give `U' and they give the same output then it is our answer.
\end{enumerate}

\begin{figure}[H]
	\includegraphics[width=13cm]{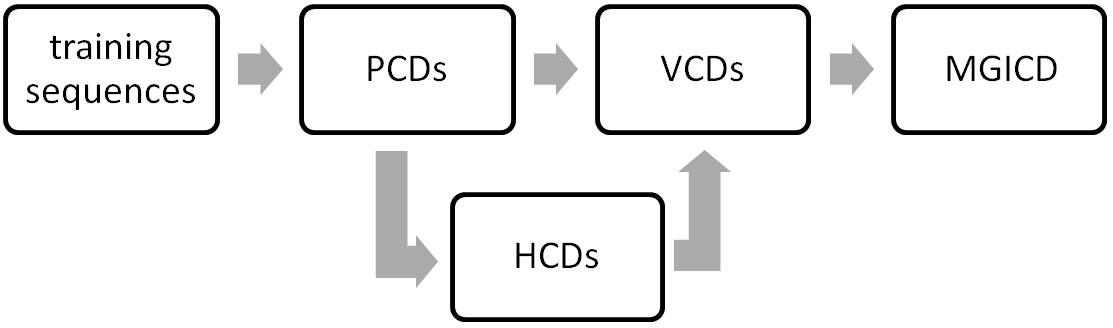}
	\centering
	\caption{Pipeline of the learning process.} \label{fig:c}
\end{figure}

Figure \ref{fig:c} shows the pipeline of the learning process starting by learning PCDs from each sequence individually, followed by merging them to build HCDs, and finally building the MGICD from the valid PCDs and HCDs.

\subsection{Cycles} \label{Cycles}
Cycle are one of the most powerful components in CDL as it has great ability in capturing invariant features and does not require adding more nodes. However, cycle learning is very computationally expensive and that is why we need to add a few restrictions while learning them. Three experiments are illustrated here to show the importance of cycles in learning and our system achieves 100\% accuracy in all of them.

\begin{table}[H]
	\begin{minipage}{1\linewidth}
		\centering
		\begin{tabular}{|l|l|}
			\hline
			Training data & Test data \\
			\hline
			1 2 3 E a b = c = d = f g =  {\color{red} f} & 	1 2 3 4 5 E = = f = f = f = y =  {\color{red} y}\\
			4 5 E b = c d = n g = t =  {\color{red} n} & h o E = 1 = 2 = 4 = 5 6 7 = 8 =  {\color{red} 2}\\
			7 E a  =  m d = f = l =  {\color{red} m} & \\
			1 2 3 4 5 6 E f s = = = = = a = g =  {\color{red} g} & \\
			\hline
		\end{tabular} 
		\caption{The model has to learn to do these steps sequentially: (1) find `E'; (2) if there are $n$ words before `E' then find the $n^{th}$ `=' counting from left; (3) the answer is the word after that sign.}\label{tab:d}		
	\end{minipage} 

\end{table}

\begin{figure}[H]
	\includegraphics[width=13cm]{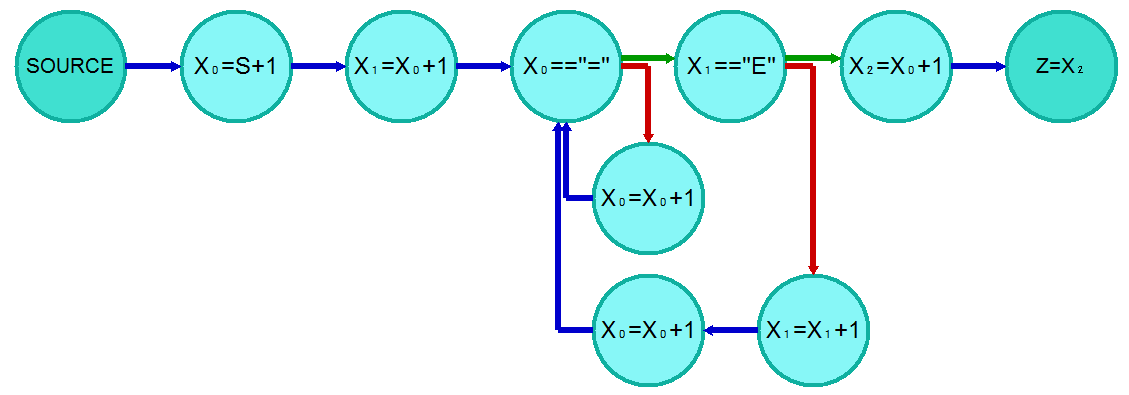}
	\centering
	\caption{CD learned from the dataset in Table \ref{tab:d}} \label{fig:d}
\end{figure}

The first dataset is in Table \ref{tab:d} and it represents a beautiful pattern that shows the capability of CDL to a learn complex pattern from only four sequences and generalize well to other sequences without overfitting their length, positions of `E', or the equal signs. Moreover, this pattern is difficult to recognize even by humans, especially if we rewrote those sequences in small alphabets only.
The final model MGICD contains one CD as shown in Figure \ref{fig:d} that has two cycles.

The second dataset, which is represented in Table \ref{tab:e}, contains three tasks: one of them is the same as the first dataset but with different letters, which makes it unreadable by humans; the other two tasks are about checking whether the  two consecutive sequences are equal or not. 

\begin{figure}[H]
	\includegraphics[width=13cm]{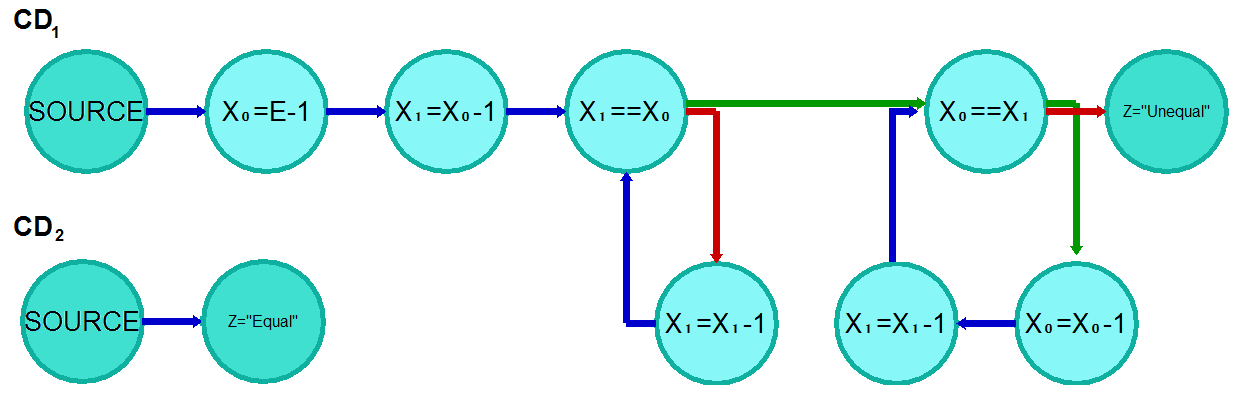}
	\centering
	\caption{CD learned from the dataset in Table \ref{tab:e}} \label{fig:e}
\end{figure}

\begin{table}[H]
	\begin{minipage}{1\linewidth}
		\centering

		\begin{tabular}{|l|l|}
				\hline
				Training data & Test data\\	
				\hline
				e h j v a b i c i d i f g i {\color{red} f}& e h j k o v i i f i f i f i y i {\color{red} y}\\
				k o v b i c d i n g i t i {\color{red} n}& h o v i 1 i 2 i 4 i 5 6 7 i 8 i {\color{red} 2}\\
				q v a i m d i f i l i {\color{red} m}& 8 9 10 11 = 8 9 10 11 = {\color{red} Equal}\\
				e h j k o p v f s i i i i i a i g i {\color{red} g}& 16 17 18 = 18 16 17 = {\color{red} Unequal}\\
				1 2 3 = 1 2 3 = {\color{red} Equal}& 16 17 18 = 16 17 16 = {\color{red} Unequal}\\
				4 5 6 7 23 24 = 4 5 6 7 23 24 = {\color{red} Equal}& 16 17 18 = 18 17 18 = {\color{red} Unequal}\\
				11 12 13 14 25 = 11 12 13 14 26 = {\color{red} Unequal}& 1 2 3 4 5 6 7 = 1 2 3 4 5 6 7 = {\color{red} Equal}\\
				40 210 = 40 210 = {\color{red} Equal}& 16 = 17 = {\color{red} Unequal}\\
				15 16 = 16 15 = {\color{red} Unequal}& \\
				17 18 = 19 18 = {\color{red} Unequal}& \\
				19 20 21 = 19 22 21 = {\color{red} Unequal}& \\
				23 24 25 = 26 24 25 = {\color{red} Unequal}& \\
				\hline
		\end{tabular} 
		\caption{A multitask learning data set. The first four sequences are the same as in Table \ref{tab:d} but in different letters. In the rest, the model also learns to decide whether two sequences are equal or not. Changing the order of sequences leads to the same learned model.}\label{tab:e}
	\end{minipage} 
	
\end{table}
The final model (MGICD) contains three VCDs, the first is positive (say CD$_0$) similar to the model in Figure \ref{fig:d} (`E' and `=' are replaced by `v' and `i'), and the other two are hybrid. One consists of CD$_2$ (in Figure \ref{fig:e}) as the positive part and CD$_1$ as the negative part, while the other consists of CD$_1$ as the positive part and CD$_0$ as the negative part.

The third dataset is to learn how to reverse sequences, but our model can only output one word. This problem is solved by re-entering the sequence in each stage. Surprisingly,it was easy to learn this task, but I noticed that during training, the CD may fail to learn the concept of reverse if the sequences before the equal sign have no repeated words. In this case, it only learns to search for the last word in that sequence and the answer is the preceding one. Therefore, I modified the dataset as shown in table \ref{tab:f} and tested that point by repeating `b' in the test sequence.

\begin{table}[!htb]
	\begin{minipage}{1\linewidth}
		\centering
		\begin{tabular}{|l|l|}
			\hline
			Training data & Test data \\
			\hline
			1 2 3 = 3 {\color{red} 2}& a b c d b f g h = {\color{red} h}\\
			4 5 4 = 4 {\color{red} 5}& a b c d b f g h = h {\color{red} g}\\
			6 7 7 = 7 7 {\color{red} 6}& a b c d b f g h = h g {\color{red} f}\\
			12 11 13 11 = {\color{red} 11}& a b c d b f g h = h g f {\color{red} b}\\
			12 11 13 11 = 11 {\color{red} 13}& a b c d b f g h = h g f b {\color{red} d}\\
			15 14 11 = 11 14 {\color{red} 15}& a b c d b f g h = h g f b d {\color{red} c}\\
			8 9 10 10 10 8 = 8 10 10 10 {\color{red} 9} & a b c d b f g h = h g f b d c {\color{red} b}\\
									                   & a b c d b f g h = h g f b d c b {\color{red} a}\\
			\hline
		\end{tabular} 
		\caption{This dataset is used to teach the CD the reverse task with different examples.}\label{tab:f}		
	\end{minipage} 
\end{table}
\begin{figure}[H]
	\includegraphics[width=13cm]{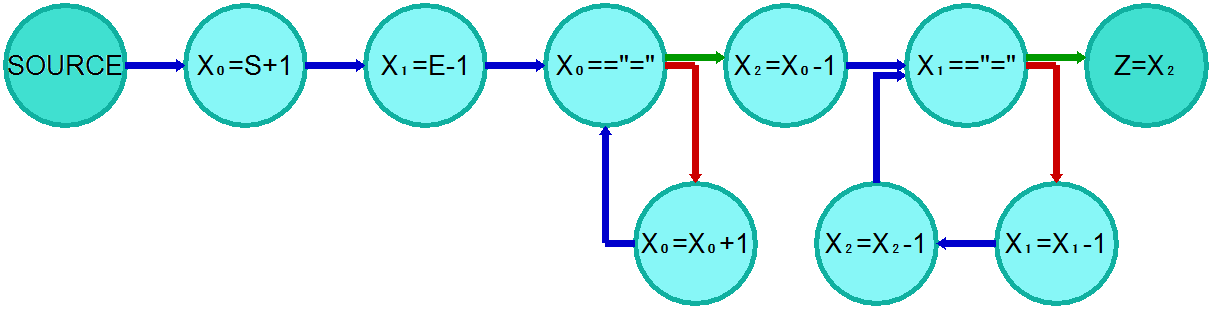}
	\centering
	\caption{CD learned from the training sequences in Table \ref{tab:f}. } \label{fig:f}
\end{figure}

\subsection{Explicit Memory}
Our knowledge can be classified into two types \citep[chap.~10]{Goodfellow:2016}: implicit  and explicit knowledge. Implicit knowledge can be learned from a number of examples and generalized to others, such as the concept of an animal. This type is what most of classical learning models are trying to learn.  Explicit knowledge  needs to be memorized rather than learned such as facts and stories.  \citet{weston:2014} talked about the need for memory to support current neural networks in question answering tasks and proposed memory networks. In CDL, we have a pretty simple approach to memorizing the facts or important sequences that we use in solving other tasks. As stated before, when CD starts to solve a question two predefined variables are automatically defined: $S$ before the first word, and $E$ after the last word. Similarly, CD considers any sequence in training data as an important memory to memorize by having two predefined variables pointing to its boundaries and defining variables over that sequence. Selecting which sequences to memorize is a learnable feature that CDL has. To demonstrate how powerful CDL is in handling memory in a very simple approach, we start by this experiment in Table \ref{tab:g}. We want to learn how to decide whether the first digit is greater than the second or not.

\begin{table}[H]
	\begin{minipage}{1\linewidth}
		\centering

		\begin{tabular}{|l|l|}
			\hline
			Training data &	Test data \\
			\hline			
			0 1 2 3 4 5 6 7 8 {\color{red} 9} & 6 \textgreater \space 1 {\color{red} Right}\\
			9 \textgreater \space 0 {\color{red} Right}& 1 \textgreater \space 6 {\color{red} Wrong}\\
			8 \textgreater \space 6 {\color{red} Right}& 1 \textgreater \space 0 {\color{red} Right}\\
			2 \textgreater \space 1 {\color{red} Right}& 7 \textgreater \space 7 {\color{red} Wrong}\\
			4 \textgreater \space 4 {\color{red} Wrong}& 0 \textgreater \space 8 {\color{red} Wrong}\\
			9 \textgreater \space 3 {\color{red} Right}&\\
			0 \textgreater \space 9 {\color{red} Wrong}&\\
			7 \textgreater \space 8 {\color{red} Wrong}&\\
			1 \textgreater \space 2 {\color{red} Wrong}&\\
			1 \textgreater \space 5 {\color{red} Wrong}&\\
			3 \textgreater \space 9 {\color{red} Wrong}&\\
			\hline
		\end{tabular} 
		\caption{The model has to learn to compare two digits using the first sequence, and decide if the first digit is greater than the second or not. }\label{tab:g}		
	\end{minipage} 
\end{table}

\begin{figure}[H]
	\includegraphics[width=13cm]{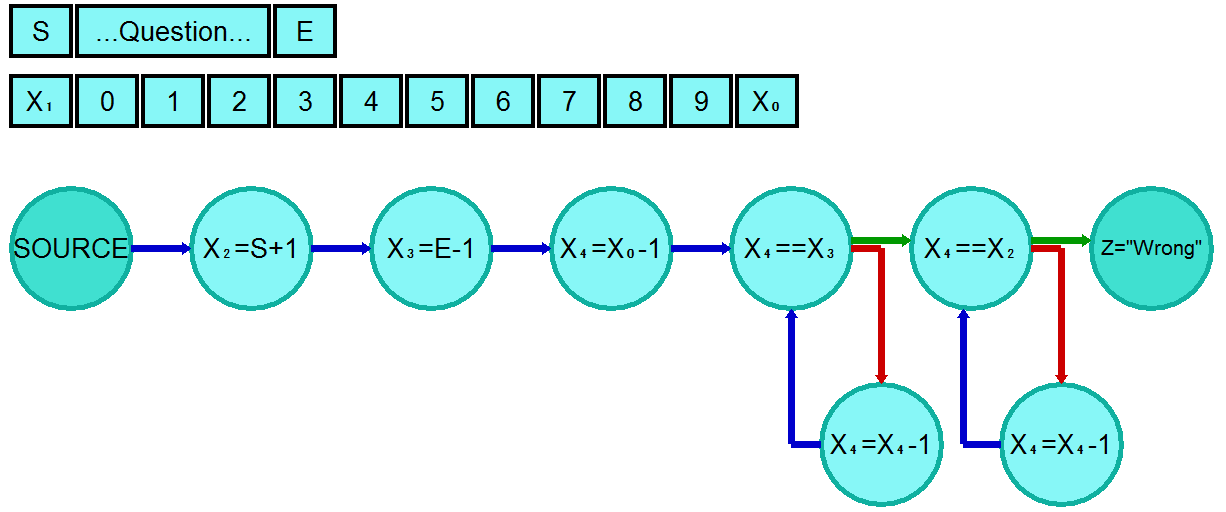}
	\centering
	\caption{This CD checks whether the comparison is wrong by defining its variables over question and memory (between $x_1$ and $x_0$).} \label{fig:g}
\end{figure}

We will not mention here the three VCDs in MGICD. Learning to answer the first sequence is not our objective, but CD was able to learn it by checking if the second word is `1', then the answer is `9'. CD in Figure \ref{fig:g} learns when the comparison is false, but notice that it cannot be used alone because it is invalid (gives `Wrong' with the first sequence). The same CD can be used as a negative description to a short CD that gives `Right'.

In the CD there are two predefined variables $x_0$ and $x_1$ that reference the first sequence in the dataset. The CD compares two digits by searching for the second digit in the sequence (memory), then searches for the first digit before it. Reaching the $x_1$ forces the CD to go to SINK node and the answer will be undefined. Otherwise, the answer is `Wrong'.

\subsection{Deep CDs} \label{DeepCD}
Deep neural networks are composed of multiple layers, each of them transforms the representation from one level to a more abstract level  \citep{lecun:2015, Goodfellow:2016}. Deep CD is a different flavor of deep models because it is not composed of different processing layers, but it has to learn: 
\begin{enumerate*}[label=\roman*)]
	\item to break down the input sequence into simpler subproblems;
	\item to solve the generated subproblem by any CD in MGICD; and
	\item to return the answer to the deep CD.
\end{enumerate*}
Here we separated the generation of subproblem (lower level representation) from solving it, which means that the deep CD only generates the subproblem but it does not solve it.

\begin{figure}[H]
	\includegraphics[width=9cm]{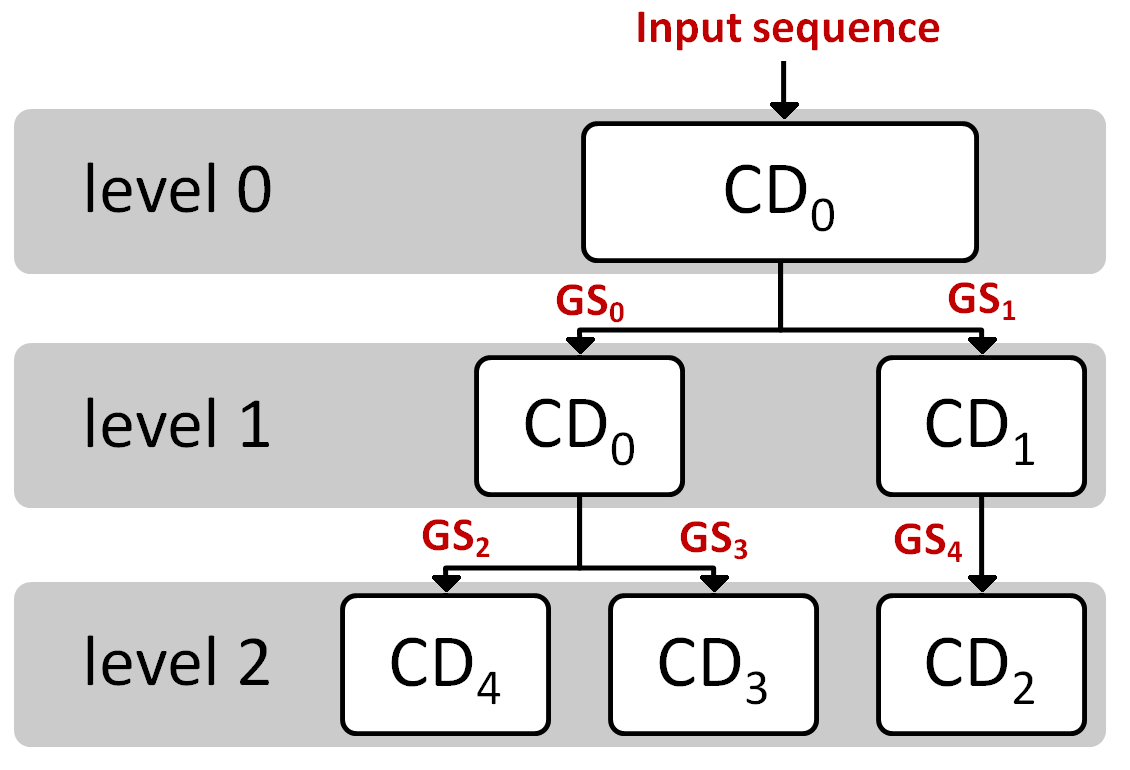}
	\centering
	\caption{This example shows how deep CDs work when we have input sequence and five VCDs in MGICD; CD$_0$ and CD$_1$ are deep models} \label{fig:j}
\end{figure}

For example, suppose we have five VCDs in a MGICD which are used to solve a new input sequence. As shown in Figure \ref{fig:j}, all CDs give `U' except CD$_0$, and thus we put it in level 0. Then it generates two subproblems $GS_0$ and $GS_1$ that are solved by CD$_0$ and CD$_1$ respectively and so on. We can say that the dynamic architecture of MGICD is constructed by its CDs and shaped by the input sequence. One important advantage of deep CDs over other learning models is that two input sequences with the same length and different words may produce completely different architectures.

In Neuropsychology, working memory can be described as a cognitive system that is responsible for manipulating information in our brain, which is important for reasoning and the guidance of decision making and behavior \citep{diamond:2013, baddeley:2012, hazy:2006}. 
A similar component in our model is the CD queue which is used to save the generated subproblem. Only two nodes are required for deep CD: push node which is used to insert a word at the end of the queue, and solve node which is used to solve the subproblem in the queue by applying MGICD on it and defining a new variable that carries the returned answer. 
The first push to occur after a solve node clears the queue before pushing its word.
Two complex datasets are used to explain how deep CDs works. `Complex' here refers to the difficulty that would face humans in solving them if numbers and equal signs were replaced by letters. 

\begin{table}[H]
	\begin{minipage}{1\linewidth}
		\centering

		\begin{tabular}{|l|l|}
			\hline
			Training data & 	Test data \\	
			\hline
			Numbers 0 1 2 3 4 5 6 7 8 9 10 11 12 13 14 15 16 17 {\color{red} 18}& 6 7 4 = {\color{red} 17}\\
			1 2 = {\color{red} 3}& 6 6 6 = {\color{red} 18}\\
			2 1 = {\color{red} 3}& 0 0 1 = {\color{red} 1}\\
			3 1 = {\color{red} 4}& 3 6 1 = {\color{red} 10}\\
			4 1 = {\color{red} 5}&\\
			3 4 = {\color{red} 7}&\\
			4 2 = {\color{red} 6}&\\
			5 1 = {\color{red} 6}&\\
			5 2 = {\color{red} 7}&\\
			3 5 = {\color{red} 8}&\\
			5 4 = {\color{red} 9}&\\
			1 2 1 = {\color{red} 4}&\\
			3 1 1 = {\color{red} 5}&\\
			1 2 5 = {\color{red} 8}&\\
			3 1 2 = {\color{red} 6}&\\
			\hline
		\end{tabular}
		\caption{This model has to learn adding two digits and then learn adding three digits by breaking them down into additions of two digits.}\label{tab:h}		
	\end{minipage} 
\end{table}

\begin{figure}[H]
	\includegraphics[width=13cm]{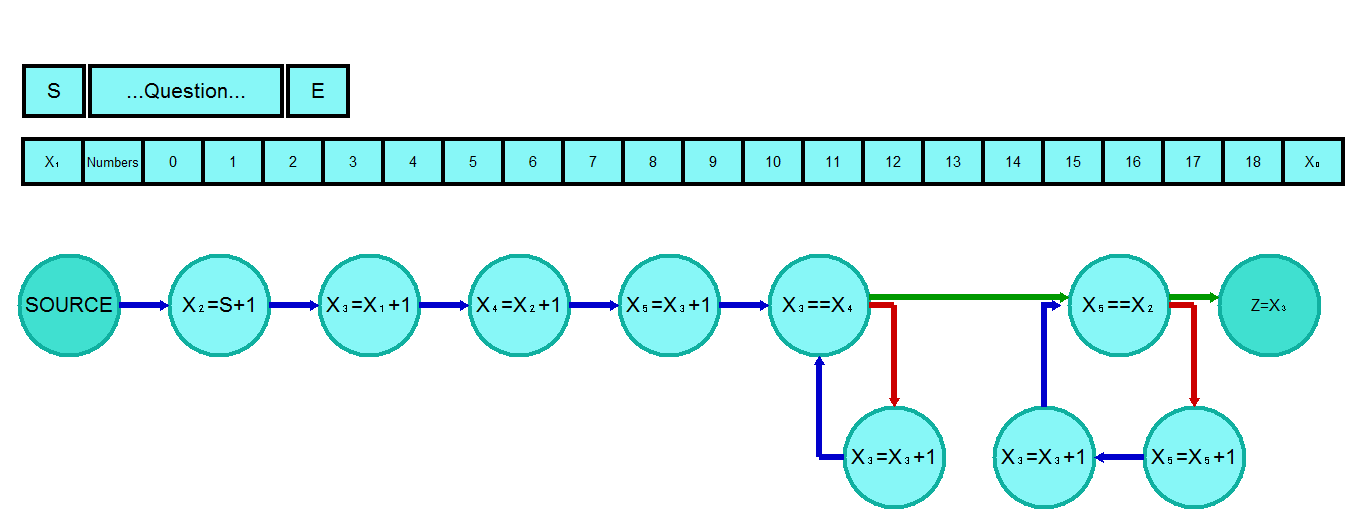}
	\centering
	\caption{This CD adds two digits and return the summation in Z variable} \label{fig:h}
\end{figure}

The first dataset in Table \ref{tab:h} includes learning the addition of two one-digit numbers and three one-digit numbers. The first task can be learned using the  first sequence as shown in Figure \ref{fig:h}. 
It simply searches for the second digit of question in the first sequence and then moves right the same number of steps required to find the first digit of the question starting from zero (invalid CD but used in HCD as positive part in the final model).

\begin{figure}[H]
	\includegraphics[width=13cm]{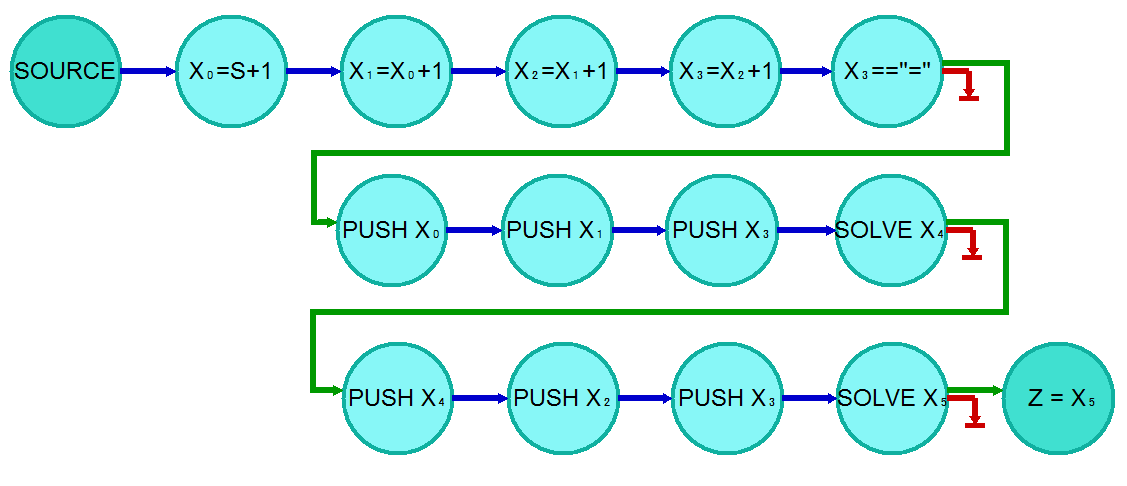}
	\centering
	\caption{This CD adds three digits in two steps making use of the learned model in Figure \ref{fig:h}. It adds the first two digits and then adds their summation to the third digit.} \label{fig:i}
\end{figure}

The second CD in Figure \ref{fig:i} is valid and solves the second task by generating two related subproblems that can be solved by the first CD. It starts by adding the first two digits and the result is used in constructing the second subproblem, and this shows how generating subproblems can be very powerful and different from other standard deep models.

\begin{figure}[H]
	\includegraphics[width=13cm]{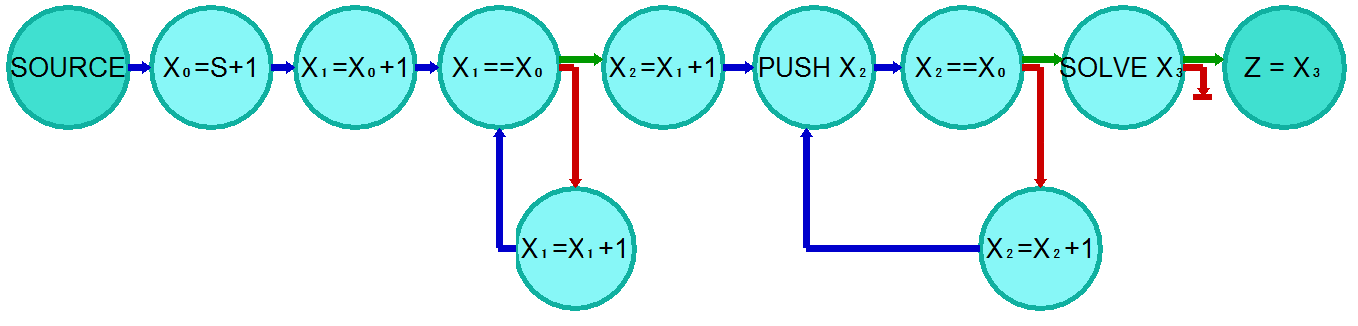}
	\centering
	\caption{Deep CD learned from dataset in Table \ref{tab:i}} \label{fig:k}
\end{figure}

The second dataset in Table \ref{tab:i} shows a very challenging problem that requires variable-depth model to be solved recursively. The MGICD that we got in this experiment contains five valid CDs. The first three sequences can be solved by one CD (the answer equals the first word). The next three have no regularity between them or other sequences, and thus they require three CDs to be solved. The remaining sequences can be solved by one deep CD that is shown in Figure \ref{fig:k}. They start with some symbol that tells you that the real question is located between two other similar symbols.

For example, in the sequence ( * d * 1 2 3 4 6 * {\color{red} 7} ), the deep CD generates the subproblem  ( 1 2 3 4 6 ) which can be solved by other CDs in the MGICD. However, in the test sequences we have two questions each of which requires different depth and ends with a different kind of problems. 

\begin{table}[H]
	\begin{minipage}{1\linewidth}
		\centering

		\begin{tabular}{|l|}
			\hline
			Training data  \\
			\hline
			1 = {\color{red} 1}\\ 
			2 = {\color{red} 2} \\
			3 = {\color{red} 3}\\
			1 2 3 4 6 {\color{red} 7}\\
			1 2 3 5 {\color{red} 8}\\
			1 2 4 {\color{red} 9}\\
			( s d x f ( 1 2 3 4 6 ( s s {\color{red} 7}\\
			\text{[ b vd [ 1 2 3 5 [ s w f {\color{red} 8}}\\
			\text{( j kd s ( 1 2 4 ( d s {\color{red} 9}}\\
			\text{* d * 1 2 3 4 6 * {\color{red} 7}}\\
			$|$ h i j k l $|$ 1 2 4 $|$ m n {\color{red} 9}\\
			$|$ $|$ 1 = $|$ {\color{red} 1}\\
			\text{M 3 b M 2 = M d d d {\color{red} 2}}\\
			\text{N c N 3 = N f f {\color{red} 3}}\\
			\text{H g d d d H 1 = H s s {\color{red} 1}}\\
			\hline			
			Test data \\
			\hline
			\{ 12 100 \{ ( h gh gg ggg ( 6 = ( ss ss \{ 1 = 100 {\color{red} 6}\\
			\{ 12 100 \{ ( h gh gg ggg ( * d * 1 2 3 4 6 * 7 ( ss ss \{ 100 = 100 {\color{red} 7}\\
			\hline
		\end{tabular} 
		\caption{The model has to learn different tasks here: the first three share the same pattern; the next three should be memorized; and the rest can be solved by one deep CD. This CD generates subproblems, which may be solvable by any of the CDs including the deep one.}\label{tab:i}		
	\end{minipage} 
\end{table}

\begin{figure}[H]
	\includegraphics[width=15cm]{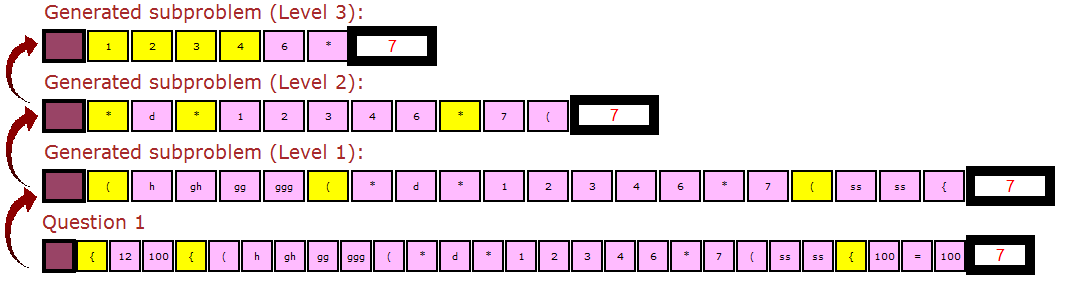}
	\centering
	\caption{Screenshot after the visualization ends that shows the steps taken by the  CD in Figure \ref{fig:k} to solve the second test sequence in Table \ref{tab:i}.} \label{fig:l}
\end{figure}

The best way to understand how this CD works is by running the visualization tool that shows the solution steps and the generated sub problems. Considering the second test sequence, Figure \ref{fig:l} is a screenshot after the visualization ends. The reader can notice that the generated sub problems always have one more word at its end but that does not affect the solution.

The domain of deep CD in Figure \ref{fig:k} may get larger if we added more VCDs learned from new tasks to the MGICD. This is because the deep CD works by extracting the problem and solving it using other VCDs in the MGICD including itself. Meaning that the new learned models can be used by the previously learned ones and vice versa. Deep CDs in general have this feature which suggests that deep CDs can make great progress in transfer learning \citep{torrey:2009, pan:2010} that attempts to develop new methods to transfer knowledge learned from one or more tasks and use it to improve learning in a related task.

\section{Learning Algorithms}
Learning the right description that captures the common patterns in data has three main stages: i) Learning PCDs, ii) Learning VCDs, and iii) Learning MGICD.
Most of the training time is always spent in the first stage in most of the experiments, and it also has big effect on the other stages. Therefore, several  ideas are developed to optimize this stage.

\subsection{Learning PCDs}
\begin{figure}[H]
	\includegraphics[width=13cm]{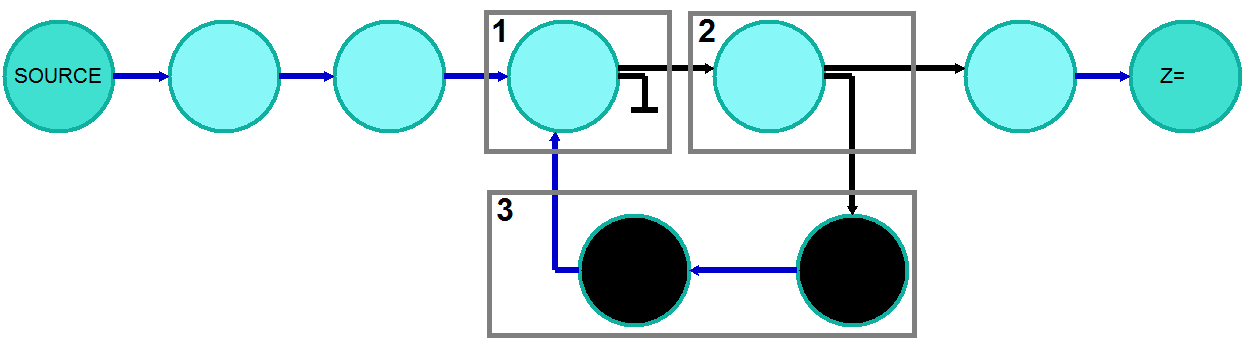}
	\centering
	\caption{This figure shows the structure of the CD where all nodes are in the main path except the black ones. It also shows the three components of the cycle: the first component is optional and can be conditional or push nodes, the second is a conditional node, and the third must be one or two assignment nodes that update the cycle variables.} \label{fig:m}
\end{figure}

CD is a planar directed graph composed of nodes and structured as a main path interrupted by cycles. The main path carries the nodes that should be passed through to move from Source to the Z node as shown in Figure \ref{fig:m}. Learning PCDs is done on each sequence in the training data separately and has two main objectives: the first is to reduce the number of repeated descriptions, and the second is to avoid producing PCDs that will not survive in the next learning stages. The characteristic vectors are used in all stages for analysis but the training data is only used in the first stage.

Algorithm \ref{PCDsLearning} is used to find all possible descriptions that satisfy these objectives. It starts with $tmpCD$ initialized with three nodes: Source, Sink and Z node, and builds over them recursively. At each time the algorithm has two decisions to make: i) whether or not to save the $tmpCD$, and ii) whether or not to continue adding nodes to the $tmpCD$.
Characteristic vector of $tmpCD$ plays an important role in making these decisions.

\noindent
\textbf{CONJECTURE 1.} If we found many PCDs with the same characteristic vectors, then the one with the least number of nodes is the one that will survive in the next learning stages.

\noindent
\textbf{CONJECTURE 2.} The number of `U' in the characteristic vector of the $tmpCD$ is monotonically increasing when adding nodes or cycles to it.

The first conjecture is based on the fact that the next stages only deal with the characteristic vectors not the CD. Consequently, the PCDs with the same vectors will have the same chance in surviving until the last stage, and if one of them is to be used in MGICD then it must be the one with the least number of nodes.

The second conjecture is based on the idea that  each time we add some node to $ tmpCD $, then its domain becomes smaller. Consequently, the number of `U' should increase, or at least do not change, because the training sequences that satisfy the description of the $ tmpCD $ are subsets of its domain. 

Based on these two conjectures the training algorithm must stop when the characteristic vector of the $tmpCD$ contains one `R' while the rest of the entries are `U's. That is because we know that the number of `U' can not decrease (conjecture 2), and if it stayed the same
then the later PCDs will have the same vector but more nodes, and  thus is not useful (conjecture 1). Also, if `R' changed, then it can not be used because we only learn a PCD with at least one `R' in its characteristic vector\footnote{In all the experiments presented in this paper, we only learn PCD that has at least one `R' in its characteristic vector and that made learning faster and did not affect the performance. However, There is no strong argument to support this. For example, if PCD has characteristic vector such as $(W,U,U,U)$, then it can be used as a negative description that rejects the first sequence.}.
Algorithm \ref{PCDsLearning} also checks whether the node is useful or not. We consider that all nodes are useful except the conditional node when its comparison is true for all training sequences, because the same CD without it will have the same characteristic vector and less nodes.

\begin{algorithm}[H]
	\caption{Learning PCDs}
	\label{PCDsLearning}
	\begin{algorithmic}[1]
		\Procedure{AddNodeCycle}{$tmpCD$}
		
		\State $fOverBuild = true$
		
		\If {\text{first call}} 
		\State initialize $tmpCD$
		\EndIf

		\Loop{\text{ on all possible Z nodes}}
		\State $p =$ \text{characteristec vector of } $tmpCD$ \text{tested on all training sequences}
		
		\If {($nRA <= 1$ {\bf And} $nWA = 0$) {\bf Or} Last node is unuseful} 
		\State $fOverBuild = false$
		\EndIf
		
		\If {$nRA > 0$ {\bf And} Last node is useful {\bf And} no saved CD with same $p$}
		\State save $tmpCD$ and $p$
		\EndIf
		
		\EndLoop
		
		\If {$fOverBuild$} 
		\Loop{\text{ on all possible nodes}}
		\State Add Node or cycle to $tmpCD$
		\State AddNodeCycle(tmpCD)
		\State Remove Node or cycle
		\EndLoop	
		\EndIf

		\EndProcedure
	\end{algorithmic}
\end{algorithm}

\noindent
\textbf{Assignment node:} In each step the algorithm tries all possible assignments. For example, in the first call we have only two variables E and S, so we have two possible assignments. If we define a variable using positive assignment, then it can not be later used in negative assignments because we already have variables defined before it. This rule is applied also for negative assignments but is not applied to a variable that has been defined in a cycle because its previous value is lost.

\noindent
\textbf{Conditional node:} In each step the algorithm tries all possible conditional nodes. It compares between two variables where the second can be constant or variable. If the condition is not in a cycle, then the false outgoing edge must be connected to the sink node because we only learn positive descriptions as explained in subsection. \ref{Positive and Negative Descriptions}. 

\noindent
\textbf{Memorized Sequences:} The algorithm can choose any sequence to memorize by having two predefined variables pointing to its boundaries and defining variables over that sequence. In all experiments, we restricted that sequence to be the first sequence in the dataset, which means the algorithm only decides whether the first sequence is needed to be memorized or not.

\noindent
\textbf{Push node:} In each step the algorithm tries all possible push nodes on the condition that the current CD queue is identical to the first part of one or more of the training sequences. The reason behind this is to reduce the number of the generated subproblems during training. However, during testing, the same CD can generate new subproblems that do not exist in the training data.

\noindent
\textbf{Solve node:} There is always one solve node to try because it has no input. But it defines a new variable carrying the answer of the subproblem.

\noindent
\textbf{Cycle:} As shown in Figure \ref{fig:m},  cycles consist of three components: the first is optional and can be conditional or push node, the second must be a conditional node, and the third  must be one or two assignment nodes that update the cycle variables. The algorithm tries all possible cycles.

\noindent
\textbf{Deep CD learning:} During learning the positive CDs, in each step we test $tmpCD$ on the training sequences, but deep CD generates a subproblem that needs to be solved by MGICD which is not learned yet. To solve this problem we must start by learning prior MGICD without push and solve nodes. Then, we learn the posterior MGICD with push and solve nodes but now we can solve the generated subproblems with the prior MGICD. The learned posterior MGICD is the final model that we use to solve both the input sequence and the generated subproblems. We added a restriction that while learning a deep CD from some training sequence, the generated subproblems can only be  solved by VCDs in prior MGICD that solves the previous sequences to that sequence. Therefore, the sequences that require deep CD to solve are always put in the last part of the dataset like the datasets in Tables \ref{tab:h} and \ref{tab:i}.

\subsection{Learning VCDs} 
VCDs are the PCDs and HCDs that have no `U' in their characteristic vectors. PCDs and their vectors have already been saved by the previous algorithm, and we need to find the possible HCDs which are valid. In our experiments, the maximum number of PCDs we got was 2276.
If we allowed HCDs with only one NCD, then the brute force search is a good algorithm with complexity $O(n^2)$.
Moreover, we do not need to apply each HCD on the training data to get its characteristic vector. Instead, we can use the vectors of its positive and negative components to calculate the resultant vector based on the three rules in Table \ref{tab:m}. These rules can be easily understood from the definition of the HCD in Section \ref{HybridCD}. For example, if we have a PCD with vector $(R,R,W,U,U)$ and a NCD with vector $(U,W,R,R,U)$, then the resultant vector of HCD is $(R,U,U,U,U)$ which is valid.

\begin{table}[H]
	\begin{minipage}{\linewidth}
		\centering
		\begin{tabular}{ccccc}
			\hline Positive CD &  & Negative CD &  & HCD \\ 
			\hline
			X  & + & U &  =  & X \\ 
			X & + & W & = & U \\ 
			X & + & R & = & U \\ 
			\hline 
		\end{tabular} 
		\caption{These rules are used to calculate the characteristic vector of a HCD without testing on training  dataset. X means that it can be `U', `R', or `W' }\label{tab:m}		
	\end{minipage}
\end{table}

\subsection{Learning MGICD} \label{learningMGICD}
After we have a set of VCDs, we need to find the MGICD, which is a set of VCDs that have two properties: their characteristic vectors add up to the all-ones vector, and the number of nodes in their VCDs is the minimum we can get. 
We do not have any assumptions about the number of VCDs, which can be thought of the number of tasks that the MGICD have learned from data. 
When a CD is used as positive in a VCD and negative in one or more VCDs, we count its nodes once and that makes our optimization problem more difficult. The brute force solution would be computationally expensive, and thus we optimized it using a very simple approach that makes it work much faster.  We start by giving a score to each VCD as follows:
$$score =\frac {\#Ones}{\#Nodes}$$
$\#Ones$ is the number of ones in the characteristic vector of the VCD. Score is defined that way based on the two criteria we have. Then, we select the VCDs that have the highest scores, and their vectors must add up to the all-ones vector without intersection.
Then, we calculate the number of their nodes ($\gamma$), which is considered as an estimate for the number of nodes in the MGICD. Finally, we search for the MGICD on the condition that the number of nodes should not exceed ($\gamma$).
In our experiments, ($\gamma$) was identical to the  $\#nodes$ of the MGICD in 26 out of 32 datasets.

\subsection{Hyperparameter tuning}
In CDL, we have eight hyperparameters but unlike most machine learning algorithms, all of them represent the maximum capacity of our models. Consequently, we can always adjust them as high as possible considering the computational cost of learning. Four of them specify the maximum number of nodes that is allowed to be in the learned PCD for each type: conditional, assignment, push, and solve. The rest specify the maximum number of cycles, maximum number of negative CDs that can be used to form a HCD, maximum number of training sequences that CD can use as a memory, and lastly a flag that can be used to allow only positive assignments. 
In future work, we could combine all these hyperparameters in one hyperparameter that indicates the maximum effort that can be spent in learning data.

\section{Results and Discussion} \label{Experimental Results}
To test our proposed model, we constructed 32 small datasets. As shown in Table \ref{tab:j}, we have successfully learned 25 datasets (Group A) with the same hyperparameter settings which were adjusted manually to provide the highest capacity required to learn all datasets.

We also have successfully learned 5 datasets (Group B) that require higher-level CDs (Table \ref{tab:k}), but we used different settings of hyperparameters for dataset 3 and 2 to make learning faster. Two of these datasets have previously been presented in subsection \ref{DeepCD}.

But as expected, CDL have failed to learn the two datasets in Table \ref{tab:l}. Nevertheless, the learned models have captured very nice patterns. Their failure is expected because they require more complex cycles than what our learning algorithm provides. For example, the first dataset requires a cycle where four variables are incremented over time but the learning algorithm only tries cycles with one or two variables.

\noindent
\textbf{Capacity and overfitting}. Models with high capacity can overfit the training sequences by memorizing their properties leading to terrible performance on test data. This is true for most learning algorithms but  not for CDL. The results in Table \ref{tab:j} show that we have successfully learned 25 different datasets that require different model capacities with the same hyperparameter settings and with different number of training sequences, yet we did not have one case of overfitting. Moreover, most datasets have more than one task that vary in their complexity.

\begin{table}[H]
	\begin{minipage}{1\linewidth}
		\centering
		\begin{tabular}{|c|c|c|c|c|c|c|c|}
			\hline 
			Dataset & \#Training & \#PCDs & \#VCDs & \#VCDs in & $\gamma$ & \#Nodes & Test \\ 
			 & sequences & &  &  MGICD &  & &  accuracy \\
			\hline 0 & 4 & 19 & 15 & 2 & 15 & 10 & 100\% \\ 
			\hline 1 & 5 & 37 & 33 & 2 & 8 & 8  & 100\% \\ 
			\hline 2 & 3 & 23 & 9 & 1 & 5 & 5 & 100\% \\ 
			\hline 3 & 3 & 23 & 9 & 1 & 4 & 4 & 100\% \\ 
			\hline 4 & 9 & 92 & 448 & 3 & 16 & 16 & 100\% \\ 
			\hline 5  & 5 & 69 & 178 & 2 & 8 & 8 & 100\% \\ 
			\hline 6 & 7 & 148 & 871 & 2 & 9 & 9 & 100\% \\ 
			\hline 7 & 3 & 26 & 13 & 1 & 7 & 7 & 100\% \\ 
			\hline 8 & 6 & 561 & 13819 & 2 & 14 & 14 & 100\% \\ 
			\hline 9  & 5 & 54 & 118 & 2 & 10 & 10 & 100\% \\ 
			\hline 10 & 4 & 80 & 167 & 1 & 6 & 6 & 100\% \\ 
			\hline 11  & 5 & 83 & 187 & 2 & 11 & 11 & 100\% \\ 
			\hline 12  & 8 & 463 & 14289 & 2 & 13 & 13 & 100\% \\ 
			\hline 13  & 6 & 451 & 7398 & 1 & 11 & 11 & 100\% \\ 
			\hline 14 & 8 & 454 & 14129 & 2 & 12 & 12 & 100\% \\ 
			\hline 15 & 4 & 75 & 131 & 1 & 11 & 11 & 100\% \\ 
			\hline 16 & 4 & 80 & 167 & 1 & 9 & 9 & 100\% \\ 
			\hline 17 & 6 & 149 & 929 & 2 & 12 & 12 & 100\% \\ 
			\hline 18 & 12 & 2276 & 66144  & 3 & 26 & 24 & 100\% \\ 
			\hline 19 & 7 & 1261 & 85377 & 1 & 11 & 11 & 100\% \\ 
			\hline 20 & 7 & 1133 & 69221 & 1 & 8 & 8 & 100\% \\ 
			\hline 21 & 8 & 75 & 396 & 3 & 20 & 15 & 100\% \\ 
			\hline 22 & 7 & 301 & 836 & 2 & 13 & 13 & 100\% \\ 
			\hline 23 & 7 & 315 & 1512 & 1 & 10 & 10 & 100\% \\ 
			\hline 24 & 11 & 770 & 29302 & 3 & 23 & 19 & 100\% \\ 
			\hline 
		\end{tabular} 
		\caption{Results of (Group A) that contains 25 different datasets such as copying, reverse, comparing two sequenecs, length concept, and counting. All trained with the same hyperparameter settings.}\label{tab:j}		
	\end{minipage}
\end{table}

\begin{table}[H]
	\begin{minipage}{1\linewidth}
		\centering

		\begin{tabular}{|c|c|c|c|c|c|c|c|}
			\hline 
			Dataset & \#Training & \#PCDs & \#VCDs & \#VCDs in & $\gamma$ & \#Nodes & Test  \\ 
			& sequences & &  &  MGICD &  & & accuracy \\ 
			\hline 0 & 16 & 51 & 19 & 11 & 69 & 69 & 100\% \\ 
			\hline 1 & 12 & 51 & 15 & 8 & 47 & 47 & 100\% \\ 
			\hline 2  & 15 & 607 & 3816 & 3 & 40 & 33 & 100\% \\ 
			\hline 3  & 15 & 406 & 546 & 5 & 41 & 41 & 100\% \\ 
			\hline 4  & 10 & 65 & 16 & 7 & 46 & 46 & 100\% \\ 
			\hline 
		\end{tabular} 
		\caption{Results of (Group B) that contains 5 datasets such as addition of one-digit numbers  and comparing the age of two persons}\label{tab:k}		
	\end{minipage}
\end{table}

\begin{table}[H]
	\begin{minipage}{1\linewidth}
		\centering

		\begin{tabular}{|c|c|c|c|c|c|c|c|}
			\hline 
			Dataset & \#Training & \#PCDs & \#VCDs & \#VCDs in & $\gamma$ & \#Nodes & Test  \\ 
			 & sequences &  &  &  MGICD &  & & accuracy \\ 
			\hline 0 & 10 & 806 & 52067 & 2 & 14 & 14 & 60\% \\ 
			\hline 1 & 5 & 174 & 781 & 2 & 22 & 18 & 66\% \\ 
			\hline 
		\end{tabular} 
		\caption{Results of (Group C) that contains 2 datasets trained with the same hyperparameter settings.}\label{tab:l}			
	\end{minipage}
\end{table}

\noindent
\textbf{Interpretable CD.} The 32 learned MGICDs are perfectly interpretable in two ways: i) any CD can be visualized using the Python tool which was used to generate the figures of most CDs in this paper; and ii) another more powerful tool that shows how the learned models solve the training dataset, and that is how we build the training sets to have more accurate models.

\noindent
\textbf{Training Time.} These experiments were performed on a personal laptop with an Intel i3 CPU 2.10 GHz and 4 GB of RAM. Surprisingly, some datasets were learned in a couple of seconds and others required over 20 minutes. But in most datasets, training time was roughly proportional to the dataset complexity. The first 25 datasets (Group A) took two hours, the next 5 datasets (Group B) took 20 minutes, and the last 2 datasets (Group C) took 3 minutes.

\section{Limitations and future work}
CD has a big disadvantage as it can not output more than one word; but that can be solved by allowing CD to have more than one Z node in the main path or in cycles. However, this makes the analysis of the characteristic vectors more difficult. 
Studying domain intersections between the VCDs can help to find better MGICDs by avoiding intersections that lead to different answers for the same question.
Future work should develop a new approach of representing VCDs of MGICD as one netwrok in such a way that one node can be shared among VCDs.  
The number of input and output connections of some part of the network can be interpreted as a frequency of this description in previously learned VCDs. In that way they can be very powerful in telling which parts of description are more likely to occur in the future, which speeds up the learning algorithms and produce more consistent VCDs.
It also needs to support adding or modifying current connections and nodes. In that way we are getting closer to the architecture and function of the neocortex in the human brain as explained in \citep{hawkins:2007}.  
CD Learning algorithms are still slow, and optimizing them will make it possible to use more complex structures of cycles and higher values for the hyperparameters that help in scaling up our model to large databases and cope with real world problems. 
Deep CDs also need better learning algorithms that allow  generating subproblems that do not exist in the training data during training.

\newpage
\section{Conclusion}
We have introduced a new framework based on some ideas including dynamic deep architecture that is shaped by the input sequence, deep CD that has the capability of breaking down the problem into simpler subproblems, positive and negative descriptions and the relation between them, characteristic vectors used in the analysis of the CDs, and finally a new simple approach to handle a selected memory that has been specified in training. 
The final model MGICD can be interpreted using a visualization tool that shows each step it takes to solve the testing sequences. This approach makes it easy to see if it learns something else that also solves the training set. In that case, we can add counter examples to the training dataset that reject what it learned in favor of what it should learn. 
Our experiments demonstrate that it was capable of learning patterns (algorithms) from small datasets, as it has successfully learned 30 multi-task datasets that vary in complexity and generalizes well to unseen data, where 25 of these datasets were learned with the same hyperparameter settings.

\section*{Acknowledgments}
I would like to thank Mohammed Khalil for proofreading the paper and Omar Barakat for useful discussions.

\section*{Appendix A} \label{AppendixA}
The 32 datasets which are used in this paper are divided into three groups A, B, and C. More details about training and results can be found at {\color{blue} \url{https://github.com/BasemElbarashy/CDL}}.

\begin{table}[H]	
	\begin{minipage}{0.5\linewidth}
		\centering
		\begin{tabular}{|l|l|}
			\hline
			Training data &	Test data  \\
			\hline
			1 {\color{red} short} & A V {\color{red} short}\\
			3 4 5 {\color{red} short} & M M M M M M {\color{red} long}\\
			10 11 12 13 14 {\color{red} long} & \\
			6 7 8 9 {\color{red} long} & \\
			\hline
		\end{tabular}
		\caption{Group A dataset 0. The model has to learn to discriminate between short and long sequences}
	\end{minipage}%
	\begin{minipage}{0.5\linewidth}
		\centering
		\begin{tabular}{|l|l|}
			\hline
			Training data & Test data\\
			\hline
			A B {\color{red} M} & A K {\color{red} M} \\
			A C {\color{red} M} & O F {\color{red} G} \\
			A D {\color{red} M} &\\ 			
			A F {\color{red} G} &\\
			H F {\color{red} G} &\\
			\hline
		\end{tabular}
		\caption{Group A dataset 1.}		
	\end{minipage}%
\end{table}
\begin{table}[H]	
	\begin{minipage}{0.5\linewidth}
		\centering
		\begin{tabular}{|l|l|}
			\hline
			Training data &	Test data  \\
			\hline
			A = {\color{red}A }& C = {\color{red}C}\\
			B = {\color{red}B }& k l m n = {\color{red}n}\\
			E F = {\color{red}F}& \\
			\hline
		\end{tabular}
		\caption{Group A dataset 2.}
	\end{minipage}%
	\begin{minipage}{0.5\linewidth}
		\centering
		\begin{tabular}{|l|l|}
			\hline
			Training data & Test data\\
			\hline
			A = {\color{red}A }& C = {\color{red}C}\\
			B = {\color{red}B }& k l m n = {\color{red}k}\\
			E F = {\color{red}E}& \\
			\hline
		\end{tabular}
		\caption{Group A dataset 3.}		
	\end{minipage}%
\end{table}
\begin{table}[H]	
	\begin{minipage}{\linewidth}
		\centering
		\begin{tabular}{|l|l|}
			\hline
			Training data &	Test data  \\
			\hline
			0 0 0 {\color{red}length=3} &	A B C {\color{red}length=3}\\
			1 2 3 {\color{red}length=3}& D E F G H I J K {\color{red}length!=3}\\
			4 5 6 {\color{red}length=3}& 4 5 6 {\color{red}length=3}\\
			7 8 9 {\color{red}length=3}& 1 3 {\color{red}length!=3}\\
			10 11 {\color{red}length!=3}&\\
			12 12 {\color{red}length!=3}&\\
			13 {\color{red}length!=3}&\\
			14 14 14 14 {\color{red}length!=3}&\\
			15 16 17 18 19 {\color{red}length!=3}&\\
			\hline
		\end{tabular}
		\caption{Group A dataset 4.}
	\end{minipage}%
\end{table}

\begin{table}[H]	
	\begin{minipage}{\linewidth}
		\centering
		\begin{tabular}{|l|l|}
			\hline
			Training data &	Test data  \\
			\hline
			1 2 3 4 5 {\color{red}true}& 5 {\color{red}true}\\
			6 7 5 {\color{red}true}& 5 5 {\color{red}true}\\
			8 5 9 {\color{red}false}& 5 5 5 5 5 5 5 5 5 6 {\color{red}false}\\
			5 5 5 10 {\color{red}false}& 1 2 3 4 5 5 5 5 5 5 6 {\color{red}false}\\
			5 1 {\color{red}false}&	6 7 8 {\color{red}false}\\
			\hline
		\end{tabular}
		\caption{Group A dataset 5. The answer is `True' when the last word is `5'}
	\end{minipage}%
\end{table}

\begin{table}[H]	
	\begin{minipage}{\linewidth}
		\centering
		\begin{tabular}{|l|l|}
			\hline
			Training data &	Test data  \\
			\hline			
			1 2 3 H 32 {\color{red}Yes} & H 1 2 3 {\color{red}Yes}\\
			4 5 H 6 7 {\color{red}Yes}& H H H {\color{red}Yes}\\
			8 H 31 {\color{red}Yes}&			1 2 H {\color{red}Yes}\\
			9 10 11 12 13 14 15 H 16 17 {\color{red}Yes}&			1 2 3 4 5 6 7 8 9 0 1 10 {\color{red}No}\\
			18 19 {\color{red}No}&\\
			20 21 22 23 {\color{red}No}&\\
			24 25 26 27 28 29 30 {\color{red}No}&\\
			\hline
		\end{tabular}
		\caption{Group A dataset 6. The answer is `Yes' if the sequence contains `H'.}
	\end{minipage}%
\end{table}

\begin{table}[H]	
	\begin{minipage}{\linewidth}
		\centering
		\begin{tabular}{|l|l|}
			\hline
			Training data &	Test data  \\
			\hline			
			1 7 5 4 H 5 6 = {\color{red}5}&			H H = {\color{red}H} \\
			9 8 H 9 = {\color{red}9}& 			14 15 17 19 21 22 H 23 24 H 25 H H = {\color{red}23}\\
			H 11 H 12 = {\color{red}11}& \\
			\hline
		\end{tabular}
		\caption{Group A dataset 7. The answer is the first word after the first `H' from left to right.}
	\end{minipage}%
\end{table}

\begin{table}[H]	
	\begin{minipage}{\linewidth}
		\centering
		\begin{tabular}{|l|l|}
			\hline
			Training data &	Test data  \\
			\hline			
			5 7 6 4 H 5 8 6 = {\color{red}5}&			H H H H = {\color{red}H}\\
			0 8 H 9 9 8 10 = {\color{red}9}&			14 15 H 17 H 19 H 21 22 H 23 24 25 = {\color{red}17}\\
			H 1 H 2 H 4 5 H = {\color{red}1}&			14 14 14 14 14 14 14 14 14 14 14 14 24 H = {\color{red}24}\\
			14 H 11 12 13 = {\color{red}11}&\\
			3 8 H = {\color{red}8}&\\
			2 4 7 6 H = {\color{red}6}&\\
			\hline
		\end{tabular}
		\caption{Group A dataset 8. The answer is the first word after the first `H' from left to right Unless this word is `=', In this case the answer is the word before that `H'.}
	\end{minipage}%
\end{table}

\begin{table}[H]	
	\begin{minipage}{\linewidth}
		\centering
		\begin{tabular}{|l|l|}
			\hline
			Training data &	Test data  \\
			\hline			
			1 2 3 4 1 5 {\color{red}Yes}&			1 1 {\color{red}Yes}\\
			6 7 6 {\color{red}Yes}&			1 2 {\color{red}No}\\
			8 9 8 10 11 12 {\color{red}Yes}&			5 6 6 6 {\color{red}No}\\
			13 14 15 16 {\color{red}No}&			7 8 8 8 8 8 8 8 8 8 8 8 7 {\color{red}Yes}\\
			17 18 {\color{red}No}&\\
			\hline
		\end{tabular}
		\caption{Group A dataset 9. The answer is `Yes' if the first word is repeated in the sequence.}
	\end{minipage}%
\end{table}

\begin{table}[H]	
	\begin{minipage}{\linewidth}
		\centering
		\begin{tabular}{|l|l|}
			\hline
			Training data &	Test data  \\
			\hline			
			1 1 1 2 3 4 = {\color{red}2}&			1 9 = {\color{red}9}\\
			1 1 5 6 = {\color{red}5}&			1 1 1 1 1 1 1 1 10 = {\color{red}10}\\
			1 1 1 1 7 8 = {\color{red}7}&\\
			1 1 1 1 1 4 3 2 7 8 = {\color{red}4}&\\
			\hline
		\end{tabular}
		\caption{Group A dataset 10.}
	\end{minipage}%
\end{table}

\begin{table}[H]	
	\begin{minipage}{\linewidth}
		\centering
		\begin{tabular}{|l|l|}
			\hline
			Training data &	Test data  \\
			\hline			
			1 1 1 = {\color{red}Yes}&			5 5 5 5 5 5 5 5 5 = {\color{red}Yes}\\
			2 2 = {\color{red}Yes}&			5 5 5 5 5 4 5 5 = {\color{red}No}\\
			3 = {\color{red}Yes}&			5 5 5 5 5 5 5 4 = {\color{red}No}\\
			1 2 1 = {\color{red}No}&			6 = {\color{red}Yes}\\
			1 1 1 3 = {\color{red}No}&\\
			\hline
		\end{tabular}
		\caption{Group A dataset 11.}
	\end{minipage}%
\end{table}

\begin{table}[H]	
	\begin{minipage}{\linewidth}
		\centering
		\begin{tabular}{|l|l|}
			\hline
			Training data &	Test data  \\
			\hline
			4 5 6 6 23 24 = 4 5 6 6 23 24 = {\color{red}Equal}&			1 2 3 1 5 6 7 = 1 2 3 1 5 6 7 = {\color{red}Equal}\\
			40 40 = 40 40 = {\color{red}Equal}&			4 5 6 4 5 6 = 4 5 6 4 5 6 = {\color{red}Equal}\\
			1 2 3 2 1 = 1 2 3 2 1 = {\color{red}Equal}&			4 5 5 4 5 5 = 4 5 5 4 5 5 = {\color{red}Equal}\\
			11 12 13 14 25 = 11 12 13 14 26 = {\color{red}Unequal}&			1 1 1 1 = 1 1 1 1 = {\color{red}Equal}\\
			15 16 = 16 15 = {\color{red}Unequal}&			1 1 2 1 1 = 1 1 3 1 1 = {\color{red}Unequal}\\
			17 18 = 19 18 = {\color{red}Unequal}&			16 = 17 = {\color{red}Unequal}\\
			19 20 21 = 19 22 21 = {\color{red}Unequal}&			8 9 10 11 = 8 9 10 11 = {\color{red}Equal}\\
			4 5 5 4 4 5 = 4 5 4 4 5 5 = {\color{red}Unequal}&			16 17 18 = 18 16 17 = {\color{red}Unequal}\\
			&			16 17 18 = 16 17 16 = {\color{red}Unequal}\\
			&			16 17 18 = 18 17 18 = {\color{red}Unequal}\\
			\hline
		\end{tabular}
		\caption{Group A dataset 12.}
	\end{minipage}%
\end{table}

\begin{table}[H]	
	\begin{minipage}{\linewidth}
		\centering
		\begin{tabular}{|l|l|}
			\hline
			Training data &	Test data  \\
			\hline
			1 2 3 = 1 4 5 = {\color{red}2}&			1 2 3 = 3 2 1 = {\color{red}1}\\
			6 7 8 9 17 18 = 6 7 10 11 = {\color{red}8}&			1 2 3 4 5 6 7 = 1 2 3 4 5 6 8 = {\color{red}7}			\\
			12 13 14 15 = 12 13 14 16 = {\color{red}15}&\\
			17 = 18 = {\color{red}17}&\\
			19 20 = 19 21 22 23 24 = {\color{red}20}&\\
			1 2 3 = 1 4 3 2 = {\color{red}2}&\\
			\hline
		\end{tabular}
		\caption{Group A dataset 13.}
	\end{minipage}%
\end{table}

\begin{table}[H]	
	\begin{minipage}{\linewidth}
		\centering
		\begin{tabular}{|l|l|}
			\hline
			Training data &	Test data  \\
			\hline
			4 5 6 6 23 24 = 24 23 6 6 5 4 = {\color{red}Reversed}&			1 2 3 1 5 6 7 = 7 6 5 1 3 2 1 = {\color{red}Reversed}\\
			40 40 = 40 40 = {\color{red}Reversed}&			4 5 6 4 5 6 = 6 5 4 6 5 4 = {\color{red}Reversed}\\
			1 2 3 2 1 = 1 2 3 2 1 = {\color{red}Reversed}&			4 5 5 4 5 5 = 5 5 4 5 5 4 = {\color{red}Reversed}\\
			11 12 13 14 25 = 26 14 13 12 11 = {\color{red}notReversed}&			1 1 1 1 = 1 1 1 1 = {\color{red}Reversed}\\
			15 16 = 15 16 = {\color{red}notReversed}&			1 1 2 1 1 = 1 1 3 1 1 = {\color{red}notReversed}\\
			17 18 = 18 19 = {\color{red}notReversed}&			16 = 17 = {\color{red}notReversed}\\
			19 20 21 = 21 22 19 = {\color{red}notReversed}&			8 9 10 11 = 11 10 9 8 = {\color{red}Reversed}\\
			4 5 5 4 4 5 = 5 5 4 4 5 4 = {\color{red}notReversed}&			16 17 18 = 18 16 17 = {\color{red}notReversed}\\
				&			16 17 18 = 16 17 16 = {\color{red}notReversed}\\
				&			16 17 18 = 18 17 18 = {\color{red}notReversed}\\
			\hline
		\end{tabular}
		\caption{Group A dataset 14.}
	\end{minipage}%
\end{table}

\begin{table}[H]
	\begin{minipage}{1\linewidth}
		\centering
		\begin{tabular}{|l|l|}
			\hline
			Training data & Test data \\
			\hline
			1 2 3 E a b = c = d = f g =  {\color{red} f} & 	1 2 3 4 5 E = = f = f = f = y =  {\color{red} y}\\
			4 5 E b = c d = n g = t =  {\color{red} n} & h o E = 1 = 2 = 4 = 5 6 7 = 8 =  {\color{red} 2}\\
			7 E a  =  m d = f = l =  {\color{red} m} & \\
			1 2 3 4 5 6 E f s = = = = = a = g =  {\color{red} g} & \\
			\hline
		\end{tabular} 
		\caption{Group A dataset 15. The model has to learn to do these steps sequentially: (1) find `E'; (2) if there are $n$ words before `E' then find the $n^{th}$ `=' counting from left; (3) the answer is the word after that sign.}
	\end{minipage} 
\end{table}

\begin{table}[H]
	\begin{minipage}{1\linewidth}
		\centering
		\begin{tabular}{|l|l|}
			\hline
			Training data & Test data \\
			\hline
			3 3 5 =  {\color{red} 5}&			1 1 1 1 1 1 1 1 1 1 1 1 1 =  {\color{red} 1}\\
			6 7 6 6 8 9 10 =  {\color{red} 8}&			1 2 1 2 1 2 3 2 2 4 2 2 5 =  {\color{red} 4}\\
			11 12 12 13 12 12 14 12 =  {\color{red} 13}&			1 2 1 2 1 2 3 2 2 2 4 2 2 5 =  {\color{red} 2}\\
			15 16 17 17 16 15 =  {\color{red} 16}&\\
			\hline
		\end{tabular} 
		\caption{Group A dataset 16. The model has to search for two consecutive words which are equal, the answer is the word after them.}
	\end{minipage} 
\end{table}

\begin{table}[H]
	\begin{minipage}{1\linewidth}
		\centering
		\begin{tabular}{|l|l|}
			\hline
			Training data & Test data \\
			\hline
			a b c d c e = {\color{red} true}&			f d d = {\color{red} false}\\
			g h d h y j = {\color{red} true}&			1 2 d 1 2 = {\color{red} false}\\
			f d f = {\color{red} true}&			1 2 f 2 1 = {\color{red} false}\\
			k l m d n = {\color{red} false}&			1 2 3 4 5 6 7 8 d 8 = {\color{red} true}\\
			o r d b q = {\color{red} false}&\\
			r f t f d u f = {\color{red} false}&\\
			\hline
		\end{tabular} 
		\caption{Group A dataset 17. The answer is `true' if the word before `d' equals the word after `d'. }
	\end{minipage} 
\end{table}

\begin{table}[H]
	\begin{minipage}{1\linewidth}
		\centering
		
		\begin{tabular}{|l|l|}
			\hline
			Training data & Test data\\	
			\hline
			e h j v a b i c i d i f g i {\color{red} f}& e h j k o v i i f i f i f i y i {\color{red} y}\\
			k o v b i c d i n g i t i {\color{red} n}& h o v i 1 i 2 i 4 i 5 6 7 i 8 i {\color{red} 2}\\
			q v a i m d i f i l i {\color{red} m}& 8 9 10 11 = 8 9 10 11 = {\color{red} Equal}\\
			e h j k o p v f s i i i i i a i g i {\color{red} g}& 16 17 18 = 18 16 17 = {\color{red} Unequal}\\
			1 2 3 = 1 2 3 = {\color{red} Equal}& 16 17 18 = 16 17 16 = {\color{red} Unequal}\\
			4 5 6 7 23 24 = 4 5 6 7 23 24 = {\color{red} Equal}& 16 17 18 = 18 17 18 = {\color{red} Unequal}\\
			11 12 13 14 25 = 11 12 13 14 26 = {\color{red} Unequal}& 1 2 3 4 5 6 7 = 1 2 3 4 5 6 7 = {\color{red} Equal}\\
			40 210 = 40 210 = {\color{red} Equal}& 16 = 17 = {\color{red} Unequal}\\
			15 16 = 16 15 = {\color{red} Unequal}& \\
			17 18 = 19 18 = {\color{red} Unequal}& \\
			19 20 21 = 19 22 21 = {\color{red} Unequal}& \\
			23 24 25 = 26 24 25 = {\color{red} Unequal}& \\
			\hline
		\end{tabular} 
		\caption{Group A dataset 18.}
	\end{minipage} 
\end{table}

\begin{table}[H]
	\begin{minipage}{1\linewidth}
		\centering
		\begin{tabular}{|l|l|}
			\hline
			Training data & Test data \\
			\hline
			1 2 3 = 3 {\color{red} 2}& a b c d b f g h = {\color{red} h}\\
			4 5 4 = 4 {\color{red} 5}& a b c d b f g h = h {\color{red} g}\\
			6 7 7 = 7 7 {\color{red} 6}& a b c d b f g h = h g {\color{red} f}\\
			12 11 13 11 = {\color{red} 11}& a b c d b f g h = h g f {\color{red} b}\\
			12 11 13 11 = 11 {\color{red} 13}& a b c d b f g h = h g f b {\color{red} d}\\
			15 14 11 = 11 14 {\color{red} 15}& a b c d b f g h = h g f b d {\color{red} c}\\
			8 9 10 10 10 8 = 8 10 10 10 {\color{red} 9} & a b c d b f g h = h g f b d c {\color{red} b}\\
			& a b c d b f g h = h g f b d c b {\color{red} a}\\
			\hline
		\end{tabular} 
		\caption{Group A dataset 19. Copying task}
	\end{minipage} 
\end{table}

\begin{table}[H]
	\begin{minipage}{1\linewidth}
		\centering
		\begin{tabular}{|l|l|}
			\hline
			Training data & Test data \\
			\hline
			1 2 3 = 1 {\color{red}2}&			a b c d b f g h = {\color{red}a}\\
			4 5 4 = 4 {\color{red}5}&			a b c d b f g h = a {\color{red}b}\\
			6 7 7 = 6 7 {\color{red}7}&			a b c d b f g h = a b {\color{red}c}\\
			12 11 13 11 = {\color{red}12}&			a b c d b f g h = a b c {\color{red}d}\\
			12 11 13 11 = 12 {\color{red}11}&			a b c d b f g h = a b c d {\color{red}b}\\
			15 14 11 = 15 {\color{red}14}&			a b c d b f g h = a b c d b {\color{red}f}\\
			8 9 10 10 10 8 = 8 9 10 10 10 {\color{red}8}&			a b c d b f g h = a b c d b f {\color{red}g}\\
			& a b c d b f g h = a b c d b f g {\color{red}h}\\
			\hline
		\end{tabular} 
		\caption{Group A dataset 20. Reverse task}
	\end{minipage} 
\end{table}

\begin{table}[H]
	\begin{minipage}{0.5\linewidth}
		\centering
		\begin{tabular}{|l|l|}
			\hline
			Training data & Test data \\
			\hline
			0 1 2 3 4 5 6 7 8 {\color{red}9}&			5 {\color{red}isDigit}\\
			1 {\color{red}isDigit}&			D {\color{red}isNotDigit}\\
			4 {\color{red}isDigit}&\\
			8 {\color{red}isDigit}&\\
			2 {\color{red}isDigit}&\\
			0 {\color{red}isDigit}&\\
			A {\color{red}isNotDigit}&\\
			C {\color{red}isNotDigit}&\\

			\hline
		\end{tabular} 
		\caption{Group A dataset 21.}
	\end{minipage} 
	\begin{minipage}{0.5\linewidth}
		\centering
		\begin{tabular}{|l|l|}
			\hline
			Training data & Test data \\
			\hline
			0 1 2 3 4 5 6 7 8 {\color{red}9}&			7 after {\color{red}6}\\
			2 after {\color{red}1}&			4 after {\color{red}3}			\\
			3 after {\color{red}2}&\\
			5 after {\color{red}4}&\\
			9 after {\color{red}8}&\\
			1 after {\color{red}0}&\\
			6 after {\color{red}5}&\\
			
			\hline
		\end{tabular} 
		\caption{Group A dataset 22.}
	\end{minipage} 
	
\end{table}

\begin{table}[H]
	\begin{minipage}{0.5\linewidth}
		\centering
		\begin{tabular}{|l|l|}
			\hline
			Training data & Test data \\
			\hline
			0 1 2 3 4 5 6 7 8 {\color{red}9}&			6 before {\color{red}7}\\
			1 before {\color{red}2}&			3 before {\color{red}4}			\\
			2 before {\color{red}3}&\\
			4 before {\color{red}5}&\\
			8 before {\color{red}9}&\\
			0 before {\color{red}1}&\\
			7 before {\color{red}8}&\\

			\hline
		\end{tabular} 
		\caption{Group A dataset 23.}
	\end{minipage} 
	\begin{minipage}{0.5\linewidth}
		\centering
		\begin{tabular}{|l|l|}
			\hline
			Training data & Test data \\
			\hline
			0 1 2 3 4 5 6 7 8 {\color{red}9}&			6 $>$ 1 {\color{red}Right}\\
			9 $>$ 0 {\color{red}Right}&			1 $>$ 6 {\color{red}Wrong}\\
			8 $>$ 6 {\color{red}Right}&			1 $>$ 0 {\color{red}Right}\\
			2 $>$ 1 {\color{red}Right}&			7 $>$ 7 {\color{red}Wrong}\\
			4 $>$ 4 {\color{red}Wrong}&			0 $>$ 8 {\color{red}Wrong}\\
			9 $>$ 3 {\color{red}Right}&\\
			0 $>$ 9 {\color{red}Wrong}&\\
			7 $>$ 8 {\color{red}Wrong}&\\
			1 $>$ 2 {\color{red}Wrong}&\\
			1 $>$ 5 {\color{red}Wrong}&\\
			3 $>$ 9 {\color{red}Wrong}&\\
			
			\hline
		\end{tabular} 
		\caption{Group A dataset 24.}
	\end{minipage} 	
\end{table}

\begin{table}[H]
	\begin{minipage}{1\linewidth}
		\centering
		
		\begin{tabular}{|l|l|}
			\hline
			Training data & 	Test data \\	
			\hline
			Numbers 0 1 2 3 4 5 6 7 8 9 10 11 12 13 14 15 16 17 {\color{red} 18}& 6 7 4 = {\color{red} 17}\\
			1 2 = {\color{red} 3}& 6 6 6 = {\color{red} 18}\\
			2 1 = {\color{red} 3}& 0 0 1 = {\color{red} 1}\\
			3 1 = {\color{red} 4}& 3 6 1 = {\color{red} 10}\\
			4 1 = {\color{red} 5}&\\
			3 4 = {\color{red} 7}&\\
			4 2 = {\color{red} 6}&\\
			5 1 = {\color{red} 6}&\\
			5 2 = {\color{red} 7}&\\
			3 5 = {\color{red} 8}&\\
			5 4 = {\color{red} 9}&\\
			1 2 1 = {\color{red} 4}&\\
			3 1 1 = {\color{red} 5}&\\
			1 2 5 = {\color{red} 8}&\\
			3 1 2 = {\color{red} 6}&\\
			\hline
		\end{tabular}
		\caption{Group B dataset 2. Addition task}	
	\end{minipage} 
\end{table}

\begin{table}[H]
	\begin{minipage}{0.5\linewidth}
		\centering
		\begin{tabular}{|l|l|}
			\hline
			Training data & Test data \\
			\hline
			A O {\color{red}1}&			L B {\color{red}NO}\\
			C O {\color{red}2}&			E B {\color{red}YES}\\
			G O {\color{red}4}&			I B {\color{red}YES}			\\
			I O {\color{red}5}&\\
			E O {\color{red}3}&\\
			E J {\color{red}4}&\\
			K X {\color{red}D}&\\
			M Y {\color{red}D}&\\
			N Z {\color{red}N}&\\
			A B {\color{red}YES}&\\
			C B {\color{red}YES}&\\
			G B {\color{red}YES}&\\
			I B {\color{red}YES}&\\
			K B {\color{red}NO}&\\
			M B {\color{red}NO}&\\
			N B {\color{red}NO}&\\
			\hline
		\end{tabular} 
		\caption{Group B dataset 0.}
	\end{minipage} 
	\begin{minipage}{0.5\linewidth}
		\centering
		\begin{tabular}{|l|}
			\hline
			Training data \\
			\hline
			Ahmed is {\color{red}16}\\
			Ali is {\color{red}16}\\
			Mona is {\color{red}14}\\
			Jack is {\color{red}14}\\
			Kareem is {\color{red}15}\\
			Dalia is {\color{red}15}\\
			Is Ahmed as old as Ali ? {\color{red}Yes}\\
			Is Ahmed as old as Mona ? {\color{red}No}\\
			Is Kareem as old as Jack ? {\color{red}No}\\
			Is Ali as old as Ahmed ? {\color{red}Yes}\\
			Is Jack as old as Mona ? {\color{red}Yes}\\
			Is Jack as old as Kareem ? {\color{red}No}\\
			\hline
			Test data\\
			\hline
			Is Ahmed as old as Dalia ? {\color{red}No}\\
			Is Kareem as old as Dalia ? {\color{red}Yes}\\
			Is Kareem as old as Mona ? {\color{red}No}	\\		
			\hline
		\end{tabular} 
		\caption{Group B dataset 1.}
	\end{minipage} 
	
\end{table}

\begin{table}[H]
	\begin{minipage}{1\linewidth}
		\centering
		
		\begin{tabular}{|l|}
			\hline
			Training data  \\
			\hline
			1 = {\color{red} 1}\\ 
			2 = {\color{red} 2} \\
			3 = {\color{red} 3}\\
			1 2 3 4 6 {\color{red} 7}\\
			1 2 3 5 {\color{red} 8}\\
			1 2 4 {\color{red} 9}\\
			( s d x f ( 1 2 3 4 6 ( s s {\color{red} 7}\\
			\text{[ b vd [ 1 2 3 5 [ s w f {\color{red} 8}}\\
			\text{( j kd s ( 1 2 4 ( d s {\color{red} 9}}\\
			\text{* d * 1 2 3 4 6 * {\color{red} 7}}\\
			$|$ h i j k l $|$ 1 2 4 $|$ m n {\color{red} 9}\\
			$|$ $|$ 1 = $|$ {\color{red} 1}\\
			\text{M 3 b M 2 = M d d d {\color{red} 2}}\\
			\text{N c N 3 = N f f {\color{red} 3}}\\
			\text{H g d d d H 1 = H s s {\color{red} 1}}\\
			\hline			
			Test data \\
			\hline
			\{ 12 100 \{ ( h gh gg ggg ( 6 = ( ss ss \{ 1 = 100 {\color{red} 6}\\
			\{ 12 100 \{ ( h gh gg ggg ( * d * 1 2 3 4 6 * 7 ( ss ss \{ 100 = 100 {\color{red} 7}\\
			\hline
		\end{tabular} 
		\caption{Group B dataset 3. The model has to learn different tasks here: the first three share the same pattern; the next three should be memorized; and the rest can be solved by one deep CD. This CD generates subproblems, which may be solvable by any of the CDs including the deep one.}		
	\end{minipage} 
\end{table}

\begin{table}[H]
	\begin{minipage}{1\linewidth}
		\centering
		\begin{tabular}{|l|l|}
			\hline
			Training data & Test data \\
			\hline
			1 2 {\color{red}3}&			7 3 {\color{red}10}\\
			1 3 {\color{red}4}&			5 6 {\color{red}11}\\
			5 2 {\color{red}7}&\\
			5 3 {\color{red}8}&\\
			3 5 {\color{red}8}&\\
			2 5 {\color{red}7}&\\
			3 1 {\color{red}4}&\\
			2 1 {\color{red}3}&\\
			3 7 {\color{red}10}&\\
			6 5 {\color{red}11}&\\
			\hline
		\end{tabular} 
		\caption{Group B dataset 4. The model has learn that swapping the first two words does not change the answer. }
	\end{minipage} 
\end{table}

\begin{table}[H]
	\begin{minipage}{1\linewidth}
		\centering
		\begin{tabular}{|l|l|}
			\hline
			Training data & Test data \\
			\hline
			1 2 = 2 1 {\color{red}Yes}&			1 2 3 4 5 6 7 8 = 2 1 4 3 6 5 8 7 {\color{red}Yes}\\
			5 6 3 4 = 6 5 4 3 {\color{red}Yes}&			1 2 3 4 5 6 0 8 = 2 1 4 3 6 5 8 7 {\color{red}No}\\
			7 8 9 10 11 12 = 8 7 10 9 12 11 {\color{red}Yes}&			a b = b a {\color{red}Yes}\\
			13 14 15 16 = 14 13 16 15 {\color{red}Yes}&			a a b a a a = a a a b a a {\color{red}Yes}\\
			17 18 = 17 18 {\color{red}No}&			a a b b a a = a a a b a a {\color{red}No}			\\
			19 20 12 22 = 19 20 19 20 {\color{red}No}&\\
			23 24 25 26 27 28 = 26 25 24 23 28 27 {\color{red}No}&\\
			19 20 30 22 = 20 19 12 22 {\color{red}No}&\\
			40 40 40 40 = 40 40 40 40 {\color{red}Yes}&\\
			a a b b a a = a a b a b a {\color{red}No}&\\

			\hline
		\end{tabular} 
		\caption{Group C dataset 0.}
	\end{minipage} 
\end{table}

\begin{table}[H]
	\begin{minipage}{1\linewidth}
		\centering
		\begin{tabular}{|l|l|}
			\hline
			Training data & Test data \\
			\hline
			a b c = d c a = {\color{red}c}&			1 2 3 4 = 5 6 7 8 9 4 = {\color{red}4}\\
			d e f k = g p h d s t = {\color{red}d}&			1 2 1 1 = 6 7 2 7 = {\color{red}2}\\
			g h i j l = m h n = {\color{red}h}&			1 2 3 4 5 = 5 6 7 = {\color{red}5}\\
			o b q r z = r s t u v = {\color{red}r}&\\
			1 2 3 11 4 5 6 = 7 8 9 10 11 12 13 14 = {\color{red}11}&\\
			
			\hline
		\end{tabular} 
		\caption{Group C dataset 1.}
	\end{minipage} 
\end{table}
\newpage
\vskip 0.2in
\bibliography{CDL}
\end{document}